\documentclass{article}

\usepackage{microtype}
\usepackage{graphicx}
\usepackage{subfigure}
\usepackage{booktabs} % for professional tables
\usepackage{amsmath}
\usepackage{xcolor}
\usepackage{array}
\usepackage{amsxtra}
\usepackage{amsfonts}
\usepackage[numbers]{natbib}
\usepackage[autolanguage]{numprint}
\usepackage{dblfloatfix}
\newcommand{\eat}[1]{} % TO MAKE LARGE BLOCKS OF TEXT INVISIBLE

\newcommand{\floor}[1]{\left\lfloor#1\right\rfloor} % floor, automatic brackets
\newcommand{\td}[2]{\if*#1\else^{#1}\fi\if*#2\else_{#2}\fi} % ^?_?, *=ignore

\newcommand{\sset}[1]{\left\{\,#1\,\right\}} % { ? }, automatic brackets
\newcommand\join\Join % normal join

\DeclareSymbolFont{txsymbolsC}{U}{txsyc}{m}{n}
\SetSymbolFont{txsymbolsC}{bold}{U}{txsyc}{bx}{n}
\DeclareFontSubstitution{U}{txsyc}{m}{n}
\DeclareMathSymbol{\ljoin}{\mathrel}{txsymbolsC}{88}
\DeclareMathSymbol{\rjoin}{\mathrel}{txsymbolsC}{89}
\newsavebox\setminusbox
\newlength\setminuslen
\newcommand\xdiag{\operatorname{diag}}
\newcommand\diag[1]{\xdiag\left(#1\right)}    % diagonal matrix
\newcolumntype{C}{>{$\displaystyle}c<{$}} % centered math
\newcolumntype{L}{>{$\displaystyle}l<{$}} % left math
\newcolumntype{R}{>{$\displaystyle}r<{$}} % right math
\newcolumntype{H}{>{\setbox0=\hbox\bgroup}c<{\egroup}@{}} % hide column

\makeatletter
\renewcommand*\env@matrix[1][*\c@MaxMatrixCols c]{%
  \hskip -\arraycolsep
  \let\@ifnextchar\new@ifnextchar
  \array{#1}}
\makeatother

\newcommand{\B}[3]{B\if*#1\else_{#1}\fi(#2,#3)} % (incomplete) beta function
\newcommand{\I}[3]{I\if*#1\else_{#1}\fi(#2,#3)} % regularized incomplete beta function
\makeatletter
\def\imod#1{\allowbreak\mkern10mu({\operator@font mod}\,\,#1)}
\makeatother

\newlength\hspaceoflen

\newcommand\vect[1]{{\mathbf{#1}}}
\newcommand\va{\vect{a}}
\newcommand\vb{\vect{b}}
\newcommand\vc{\vect{c}}
\newcommand\vd{\vect{d}}
\newcommand\ve{\vect{e}}
\newcommand\vf{\vect{f}}
\newcommand\vg{\vect{g}}
\newcommand\vh{\vect{h}}
\newcommand\vi{\vect{i}}
\newcommand\vj{\vect{j}}
\newcommand\vk{\vect{k}}
\newcommand\vl{\vect{l}}
\newcommand\vm{\vect{m}}
\newcommand\vn{\vect{n}}
\newcommand\vo{\vect{o}}
\newcommand\vp{\vect{p}}
\newcommand\vq{\vect{q}}
\newcommand\vr{\vect{r}}
\newcommand\vs{\vect{s}}
\newcommand\vt{\vect{t}}
\newcommand\vu{\vect{u}}
\newcommand\vv{\vect{v}}
\newcommand\vw{\vect{w}}
\newcommand\vx{\vect{x}}
\newcommand\vy{\vect{y}}
\newcommand\vz{\vect{z}}
\newcommand\vzero{\vect{0}}

\newcommand\vtheta{\vect{\theta}}

\newcommand\mA{\vect{A}}
\newcommand\mB{\vect{B}}
\newcommand\mC{\vect{C}} 
\newcommand\mD{\vect{D}}
\newcommand\mE{\vect{E}}
\newcommand\mF{\vect{F}}
\newcommand\mG{\vect{G}}
\newcommand\mH{\vect{H}}
\newcommand\mI{\vect{I}}
\newcommand\mJ{\vect{J}}
\newcommand\mK{\vect{K}}
\newcommand\mL{\vect{L}}
\newcommand\mM{\vect{M}}
\newcommand\mN{\vect{N}} 
\newcommand\mO{\vect{O}}
\newcommand\mP{\vect{P}}
\newcommand\mQ{\vect{Q}} 
\newcommand\mR{\vect{R}} 
\newcommand\mS{\vect{S}}
\newcommand\mT{\vect{T}}
\newcommand\mU{\vect{U}}
\newcommand\mV{\vect{V}}
\newcommand\mW{\vect{W}}
\newcommand\mX{\vect{X}}
\newcommand\mY{\vect{Y}}
\newcommand\mZ{\vect{Z}}

\newcommand\bN{\mathbb{N}} % set of positive integers

\newcommand\bR{\mathbb{R}} % set of reals

\DeclareMathAlphabet{\mathcal}{OMS}{cmsy}{m}{n}

\newcommand\cE{\mathcal{E}}

\newcommand\cK{\mathcal{K}}
\newcommand\cL{\mathcal{L}}

\newcommand\cR{\mathcal{R}}

\DeclareMathAlphabet\mathbfcal{OMS}{cmsy}{b}{n}
\newcommand\tA{\mathbfcal{A}}

\newcommand\tD{\mathbfcal{D}}

\newcommand\tG{\mathbfcal{G}}
\newcommand\tH{\mathbfcal{H}}

\newcommand\tS{\mathbfcal{S}}

\newcommand\tX{\mathbfcal{X}}

\accentedsymbol\Abar{{\bar A}}
\accentedsymbol\Bbar{{\bar B}}
\accentedsymbol\Cbar{{\bar C}}
\accentedsymbol\Dbar{{\bar D}}
\accentedsymbol\Ebar{{\bar E}}
\accentedsymbol\Fbar{{\bar F}}
\accentedsymbol\Gbar{{\bar G}}
\accentedsymbol\Hbar{{\bar H}}
\accentedsymbol\Ibar{{\bar I}}
\accentedsymbol\Jbar{{\bar J}}
\accentedsymbol\Kbar{{\bar K}}
\accentedsymbol\Lbar{{\bar L}}
\accentedsymbol\Mbar{{\bar M}}
\accentedsymbol\Nbar{{\bar N}}
\accentedsymbol\Obar{{\bar O}}
\accentedsymbol\Pbar{{\bar P}}
\accentedsymbol\Qbar{{\bar Q}}
\accentedsymbol\Rbar{{\bar R}}
\accentedsymbol\Sbar{{\bar S}}
\accentedsymbol\Tbar{{\bar T}}
\accentedsymbol\Ubar{{\bar U}}
\accentedsymbol\Vbar{{\bar V}}
\accentedsymbol\Wbar{{\bar W}}
\accentedsymbol\Xbar{{\bar X}}
\accentedsymbol\Ybar{{\bar Y}}
\accentedsymbol\Zbar{{\bar Z}}

\accentedsymbol\abar{{\bar a}}
\accentedsymbol\bbar{{\bar b}}
\accentedsymbol\cbar{{\bar c}}
\accentedsymbol\dbar{{\bar d}}
\accentedsymbol\ebar{{\bar e}}
\accentedsymbol\fbar{{\bar f}}
\accentedsymbol\gbar{{\bar g}}
\makeatletter
\@ifundefined{hbar}{}{
        \let\hbar\@undefined
}
\makeatother
\accentedsymbol\hbar{{\bar h}}
\accentedsymbol\ibar{{\bar i}}
\accentedsymbol\jbar{{\bar j}}
\accentedsymbol\kbar{{\bar k}}
\accentedsymbol\lbar{{\bar l}}
\accentedsymbol\mbar{{\bar m}}
\accentedsymbol\nbar{{\bar n}}
\makeatletter
\@ifundefined{obar}{}{
        \let\obar\@undefined      
}
\makeatother
\accentedsymbol{\obar}{{\bar o}}        
\accentedsymbol\pbar{{\bar p}}
\accentedsymbol\qbar{{\bar q}}
\accentedsymbol\rbar{{\bar r}}
\accentedsymbol\sbar{{\bar s}}
\accentedsymbol\tbar{{\bar t}}
\accentedsymbol\ubar{{\bar u}}
\accentedsymbol\vbar{{\bar v}}
\accentedsymbol\wbar{{\bar w}}
\accentedsymbol\xbar{{\bar x}}
\accentedsymbol\ybar{{\bar y}}
\accentedsymbol\zbar{{\bar z}}

\accentedsymbol\mAhat{{\hat\mA}}
\accentedsymbol\mBhat{{\hat\mB}}
\accentedsymbol\mChat{{\hat\mC}}
\accentedsymbol\mDhat{{\hat\mD}}
\accentedsymbol\mEhat{{\hat\mE}}
\accentedsymbol\mFhat{{\hat\mF}}
\accentedsymbol\mGhat{{\hat\mG}}
\accentedsymbol\mHhat{{\hat\mH}}
\accentedsymbol\mIhat{{\hat\mI}}
\accentedsymbol\mJhat{{\hat\mJ}}
\accentedsymbol\mKhat{{\hat\mK}}
\accentedsymbol\mLhat{{\hat\mL}}
\accentedsymbol\mMhat{{\hat\mM}}
\accentedsymbol\mNhat{{\hat\mN}}
\accentedsymbol\mOhat{{\hat\mO}}
\accentedsymbol\mPhat{{\hat\mP}}
\accentedsymbol\mQhat{{\hat\mQ}}
\accentedsymbol\mRhat{{\hat\mR}}
\accentedsymbol\mShat{{\hat\mS}}
\accentedsymbol\mThat{{\hat\mT}}
\accentedsymbol\mUhat{{\hat\mU}}
\accentedsymbol\mVhat{{\hat\mV}}
\accentedsymbol\mWhat{{\hat\mW}}
\accentedsymbol\mXhat{{\hat\mX}}
\accentedsymbol\mYhat{{\hat\mY}}
\accentedsymbol\mZhat{{\hat\mZ}}

\accentedsymbol\vahat{{\hat\va}}
\accentedsymbol\vbhat{{\hat\vb}}
\accentedsymbol\vchat{{\hat\vc}}
\accentedsymbol\vdhat{{\hat\vd}}
\accentedsymbol\vehat{{\hat\ve}}
\accentedsymbol\vfhat{{\hat\vf}}
\accentedsymbol\vghat{{\hat\vg}}
\accentedsymbol\vhhat{{\hat\vh}}
\accentedsymbol\vihat{{\hat\vi}}
\accentedsymbol\vjhat{{\hat\vj}}
\accentedsymbol\vkhat{{\hat\vk}}
\accentedsymbol\vlhat{{\hat\vl}}
\accentedsymbol\vmhat{{\hat\vm}}
\accentedsymbol\vnhat{{\hat\vn}}
\accentedsymbol\vohat{{\hat\vo}}
\accentedsymbol\vphat{{\hat\vp}}
\accentedsymbol\vqhat{{\hat\vq}}
\accentedsymbol\vrhat{{\hat\vr}}
\accentedsymbol\vshat{{\hat\vs}}
\accentedsymbol\vthat{{\hat\vt}}
\accentedsymbol\vuhat{{\hat\vu}}
\accentedsymbol\vvhat{{\hat\vv}}
\accentedsymbol\vwhat{{\hat\vw}}
\accentedsymbol\vxhat{{\hat\vx}}
\accentedsymbol\vyhat{{\hat\vy}}
\accentedsymbol\vzhat{{\hat\vz}}

\usepackage{multicol, latexsym, amssymb}
\usepackage{subfigure}
\usepackage{hyperref}

\usepackage[accepted]{icml2019}
\newcommand\Real{\operatorname{Re}}
\newcommand{\nex}{\text{e-}}

\usepackage{mathptmx}   
\icmltitlerunning{A Relational Tucker Decomposition for Multi-Relational Link Prediction}

\begin{document}

\twocolumn[
\icmltitle{A Relational Tucker Decomposition for Multi-Relational Link Prediction}



\icmlsetsymbol{equal}{*}

\begin{icmlauthorlist}
\icmlauthor{Yanjie Wang}{to}
\icmlauthor{Samuel Broscheit}{to}
\icmlauthor{Rainer Gemulla}{to}
\end{icmlauthorlist}

\icmlaffiliation{to}{University of Mannheim, Germany}
% \icmlaffiliation{goo}{Googol ShallowMind, New London, Michigan, USA}
% \icmlaffiliation{ed}{School of Computation, University of Edenborrow, Edenborrow, United Kingdom}

\icmlcorrespondingauthor{}{ywang,rgemulla@uni-mannheim.de}
\icmlcorrespondingauthor{}{broscheit@informatik.uni-mannheim.de}


\vskip 0.3in
]



\printAffiliationsAndNotice{}  % leave blank if no need to mention equal contribution

\begin{abstract}
% Embedding approaches
% recently received  attention for modeling multi-relational data such as knowledge 
% graphs. 
  We propose the Relational Tucker3 (RT) decomposition for multi-relational link
  prediction in knowledge graphs. We show that many existing knowledge graph
  embedding models are special cases of the RT decomposition with certain
  predefined sparsity patterns in its components. In contrast to these prior
  models, RT decouples the sizes of entity and relation embeddings, allows
  parameter sharing across relations, and does not make use of a predefined
  sparsity pattern. We use the RT decomposition as a tool to explore whether it
  is possible and beneficial to automatically learn sparsity patterns, and
  whether dense models can outperform sparse models (using the same number of
  parameters). Our experiments indicate that---depending on the dataset--both
  questions can be answered affirmatively.
% In the RT view, the allowed decisions corresponds to the distinct sparsity patterns
% is the shared feature of 
% on the core tensors. 
% BM differ in which interactions are allowed  and the decisions are made a priori. We explore learning suitable allowed interactions based on data via sparse RT (SRT). Our experiment results show that decoupling entity and relation embedding sizes, parameter sharing between relations, and learning allowed interactions can all be beneficial.
% of the the better performing bilinear models,
% Our experiments show that SRT can be competitive 
% for multi-relational link prediction, but sparsity is not always necessary for achieving competitive results. 
\end{abstract}

%!TEX root = main.tex
\section{Introduction}

Knowledge graphs (KG)~\cite{LehmannIJJKMHMK15,MahdisoltaniBS15}
% such as DBPedia~\cite{LehmannIJJKMHMK15}, YAGO~\cite{MahdisoltaniBS15} and Freebase~\cite{BollackerEPST08} 
% provide valuable information for question answering~\cite{questionanwsering,QANLP2}, semantic search~\cite{SemanticSearch}, and recommendation~\cite{RECKG,RECKG2}. They 
represent facts as subject-relation-object triples, e.g., \textit{(London,
  capital\_of, UK)}. KG embedding (KGE) models embed each entity and each
relation of a given KG into a latent semantic space such that important
structure of the KG is retained. A large number of KGE models has been proposed
in the literature; applications include question answering~\cite{questionanwsering,QANLP2}, semantic search~\cite{SemanticSearch}, and recommendation~\cite{RECKG,RECKG2}.

Many of the available KGE models can be expressed as \emph{bilinear models}, on
which we focus throughout. Examples include RESCAL~\cite{NickelTK11},
DistMult~\cite{Tucker1966}, ComplEx~\cite{TrouillonWRGB16}, Analogy~\cite{Analogy}, and
CP~\cite{Canonical}. KGE models assign a ``score'' to each
subject-relation-object triple; high-scoring triples are considered more likely
to be true. In bilinear models, the score is computed using a relation-specific
linear combination of the pairwise interactions of the embeddings of the subject
and the 
object. % For a \textit{fixed} entity embedding dimension, the most general bilinear model is
% Since these models restrict the allowed interactions, we refer to them as \textit{constrained bilinear models}. 
The models differ in the kind of interactions that are considered: RESCAL is
dense in that it considers all pairwise interactions, whereas all other of the
aforementioned models are sparse in that they consider only a small, hard-coded
subset of interactions (and learn weights only for this subset). As a
consequence, these later models have fewer parameters. They empirically show
state-of-the-art performance~\cite{LiuWY17,TrouillonWRGB16,Canonical} for
multi-relational link prediction tasks.

In this paper, we propose the Relational Tucker3 (RT) decomposition, which
tailors the standard Tucker3 decomposition~\cite{Tucker1966} to the relational domain.
The RT decomposition is inspired by RESCAL, which specialized the Tucker2
decomposition in a similar way. We use the RT decomposition as a tool to to
explore (1) whether we can automatically learn which interactions should be
considered instead of using hard-coded sparsity patterns, (2) whether and when
this is beneficial, and finally (3) whether sparsity is indeed necessary to
learn good representations.

In a nutshell, RT decomposes the KG into an entity embedding matrix, a relation
embedding matrix, and a core tensor. We show that all existing bilinear models
are special cases of RT under different viewpoints: the fixed core tensor view
and the constrained core tensor view. In both cases, the differences between
different bilinear models are reflected in different (fixed a priori) sparsity
patterns of the associated core tensor. In contrast to bilinear models, RT offers a
natural way to decouple entity and relation embedding sizes and allows parameter
sharing across relations. These properties allow us to learn state-of-the-art
dense representations for KGs. Moreover, to study the questions raised above, we
propose and explore a sparse RT decomposition, in which the core tensor is
encouraged to be sparse, but without using a predefined sparsity pattern.

We conducted an experimental study on common benchmark datasets to gain insight
into the dense and sparse RT decompositions and to compare them with
state-of-the-art models. Our results indicate that dense RT models can
outperform state-of-the-art sparse models (when using the same number of
parameters), and that it is possible and sometimes beneficial to learn sparsity
patterns via a sparse RT model. We found that the best-performing method is
dataset-dependent.

% Sec.~\ref{sec:bilinear} provides background on bilinear models and the multi-relational link prediction task. The RT views of bilinear models is described in Sec.~\ref{sec:rtucker}, together with the introduction of SRT. Sec ~\ref{sec:rtucker}. The experiment setting and results are described in Sec.~\ref{sec:experiments}. Conclusions are drawn in Sec.~\ref{sec:conclusion}.  

% Given these observation, it is natural to ask (1) Do extra flexibilities inspired from the Tucker decomposition interpretation help for modeling relational data? (2) How much do the specific patterns from existing models contribute to the performance boost? and 
% (3) What are the strengths and weaknesses of this more general Tucker3 decomposition approach for relational tasks compared with existing constrained bilinear models? 
% We answer these questions based on the empirical results.

%%% Local Variables:
%%% mode: latex 
%%% TeX-master: "main"
%%% End:

%!TEX root = main.tex
\section{Background}\label{sec:bilinear}

\paragraph{Multi-relational link prediction.} Given a set of entities $\cE$ and
a set of relations $\cR$, a knowledge graph $\cK\subseteq \cE\times\cR\times\cE$ is a set of
triples $(i, k, j)$, where $i, j \in \cE$ and $k \in \cR$. Commonly, $i$, $k$ and
$j$ are referred to as the \emph{subject}, \emph{relation}, and \emph{object},
respectively. A knowledge base can be viewed as a labeled graph, where each
vertex corresponds to an entity, each label to a relation, and each labeled edge
to a triple. The goal of multi-relational link prediction is to determine
correct but unobserved triples $t'\in (\cE\times\cR\times\cE) \backslash \cK$ based on $\cK$. The
task has been studied extensively in the literature~\cite{Nickel0TG16}. The main
approaches include rule-based methods~\cite{PATH,AMIE,meilicke2018fine}, knowledge
graph
embeddings~\cite{BordesUGWY13,TrouillonWRGB16,NickelTK11,NickelRP16,Analogy,dettmers2018conve},
and combined methods such as~\cite{COMRULEEMBED}.

\paragraph{KG embedding (KGE) models.}

A KGE model associates with each entity $i$ and relation $k$ an
\textit{embedding} $\ve_i\in\bR^{d_e}$ and $\vr_k\in\bR^{d_r}$ in a low-dimensional
vector space, respectively. Here $d_e,d_r\in\bN^+$ are hyper-parameters that refer
to the \emph{size} of the entity embeddings and relation embeddings,
respectively. Each model uses a \textit{scoring function} $s:\cE \times \cR \times \cE \rightarrow
\mathbb{R}$ to associate a score $s(i,k,j)$ to each triple $(i,k,j)\in
\cE\times\cR\times\cE$. The scoring function depends on $i$, $k$, and $j$ only through
their respective embeddings $\ve_i$, $\vr_k$, and $\ve_j$. Triples with high
scores are considered more likely to be true than triples with low scores.
% Roughly speaking, the models try to find embeddings that capture the structure of the entire knowledge graph well. 
% Since embeddings constitute a form of compression, the models are forced to
% generalize so that new facts can be predicted.
% A \emph{score-based relational embedding model} associates a \emph{score}
% $s(i,k,j) \in \mathbb{R}$ with each subject-relation-object triple, which indicates the truthfulness of the triple $(i,k,j)$. 
% In the rest, operator $\circ$ stands for element-wise multiplication of matrices or tensors with the same shape;
% $[n]$ is the shorthand for the set $\{1,2,\dots,n\}$; 
% %The symbol $\otimes$ stands for Kronecker product and 
% $\text{vec}(\cdot)$ denotes the vectorization of a matrix from its columns;
% $\diag{\cdot}$ refers to a block-diagonal matrix built from the arguments. By
% convention, vectors $\ve_i$ refer to rows of matrix $\mE$ (as a column vector)
% and scalars $e_{ij}$ to individual entries.

Embedding models roughly can be classified into translation-based
models~\cite{BordesUGWY13,WangZFC14}, factorization
models~\cite{TrouillonN17,Analogy}, and neural
models~\cite{SocherCMN13,dettmers2018conve}.
% Modeling relational data with tensor decomposition models have been explored extensively, especially in knowledge graphs. 
% A survey on this can be found in~\citet{Nickel0TG16}.
% Recently,~\citet{Canonical} applied CP decomposition to model knowledge graph data. In their work, an entity has different representations acting as subject and object. 
% Beside this, it differs from our work in that sparsity pattern is not learned rather fixed. 
% In contrast to our focus on bilinear models, ~\citet{dettmers2018conve} showed that modeling the interactions with neural networks can potentially improve the parameter efficiency. 
Many of the available KGE models can be expressed as bilinear
models~\cite{Bilinear}, in which the scoring function takes form
\begin{equation}\label{eq:bilinearproduct}
s(i,k,j)=\ve_i^T\mM_k\ve_j, 
% =\text{vec}(\mM_k)^T(\ve_j\otimes\ve_i),
\end{equation} 
where $\ve_i,\ve_j\in\bR^{d_e}$ and $\mM_k\in\bR^{d_e\times d_e}$. We refer to matrix
$\mM_k$ as the \textit{mixing matrix} for relation $k$; $\mM_k$ is derived from
$\vr_k$ using a model-specific mapping $\bR^{d_r}\to\bR^{d_e\times d_e}$. Existing
bilinear models differ from each other mainly in this mapping. We summarize some
of the most prevalent models in what follows. We use $\circ$ for the Hadamard
product (i.e., elementwise multiplication), $\operatorname{vec}(\cdot)$ for the
vectorization of a matrix from its columns, $\mI_K$ for the $K\times K$ identity
matrix, $\diag{\cdot}$ for the diagonal matrix built from the arguments, and $[n]$
for $\{1,2,\dots,n\}$. By convention, vectors of form $\va_i$ refer to rows of
some matrix $\mA$ (as a column vector) and scalars $a_{ij}$ to individual
entries.

\paragraph{\textbf{RESCAL}~\cite{NickelTK11}.} RESCAL is an unconstrained
bilinear model and directly learns the mixing matrices $\sset{\mM_k}$. In our
notation, RESCAL sets $d_r=d_e^2$ and uses
\[
  \mM_k^{\text{RESCAL}}=\operatorname{vec}^{-1}(\vr_k).
\]
All of the
bilinear models discussed below can be seen as constrained variants of RESCAL;
constraints are used to facilitate learning and reduce the number of parameters.

\paragraph{\textbf{DistMult}~\protect\cite{YangYHGD14a}.} DistMult puts a
diagonality constraint on the mixing matrices. The relation embeddings $\vr_k$
hold the values on the diagonal, i.e., $d_r=d_e$ and
\[
  \mM^{\text{DistMult}}_k = \diag{\vr_k}.
\]
Since each mixing matrix is symmetric, we have $s(i,k,j)=s(j,k,i)$ so that
DistMult can only model symmetric relations. DistMult is equivalent to the
INDSCAL tensor decomposition~\cite{Carroll1970}.

\paragraph{\textbf{CP}~\protect\cite{Canonical}.} CP is another classical tensor
decomposition~\cite{Tensor} and has recently shown good results for KGE. Here CP
associates two embeddings $\ve_i^{\text{sub}}$ and $\ve_i^{\text{obj}}$ with
each entity and uses scoring function of form
$s(i,k,j)=(\ve_i^{\text{sub}})^T\diag{\vr_k}\ve_j^{\text{obj}}$. The CP
decomposition can be expressed as a bilinear model using mixing matrix
\[
  \mM^{\text{CP}}_k =
  \begin{pmatrix}
    \vzero_{\frac{d_e}2 \times \frac{d_e}2} & \diag{\vr_k} \\
    \vzero_{\frac{d_e}2 \times \frac{d_e}2} & \vzero_{\frac{d_e}2 \times \frac{d_e}2}.
  \end{pmatrix},
\]
where $d_e$ is even, $d_r=d_e/2$, and thus $\vr_k\in\bR^{d_e/2}$. To see this,
observe that if we set $\ve=
\begin{pmatrix}
\ve_i^{\text{sub}} \\  \ve_i^{\text{obj}}
\end{pmatrix}$, then
$\ve_i^T\mM^{\text{CP}}_k\ve_j=(\ve_i^{\text{sub}})^T\diag{\vr_k}\ve_j^{\text{obj}}$.
Note that CP can model symmetric and asymmetric relations.

\paragraph{\textbf{ComplEx}~\cite{TrouillonWRGB16} 
% and \textbf{HolE}~\protect\cite{NickelRP16}
.}  
% A ComplEx model with $\ve_i\in \mathbb{C}^{n}$ can be re-parameterized with $\ve_i\in\mathbb{R}^{2n}$. In other words, for real space parameterization, 
% the embedding size $d_e=d_r$ must be even. 
% Using real space parameterization, 
ComplEx is currently one of the best-performing KGE models (see also
Sec.~\ref{sec:compvsprior}). Let $d_e$ be even, set $d_r=d_e$, and denote by
$\vr_k^{\text{left}},$ and $\vr_k^{\text{right}}$ the first and last $d_e/2$
entries of $\vr_k$. ComplEx then uses mixing matrix
\[
  \mM^{\text{ComplEx}}_k=\begin{bmatrix} \nonumber
    \text{diag}(\vr_k^{\text{left}}) & \text{diag}(\vr_k^{\text{right}}) \\
    -\text{diag}(\vr_k^{\text{right}}) & \text{diag}(\vr_k^{\text{left}})
  \end{bmatrix}.
\]
As CP, ComplEx can model both symmetric ($\vr_k^{\text{right}}= \mathbf{0}$) and
asymmetric ($\vr_k^{\text{right}}\neq \mathbf{0}$)
relations. % ComplEx was proposed to address the issue of modeling asymmetric relations with DistMult.

ComplEx can be expressed in a number of equivalent
ways~\cite{SimpleE}. In their original work,
\citet{TrouillonWRGB16} use complex embeddings (instead of real ones) and
scoring function $ s(i,k,j) = \Real(\ve_i^T\diag{\vr_k}\ve_j)$, where $\Real(\cdot)$
extracts the real part of a complex number. 
Likewise, HolE~\cite{NickelRP16} is
equivalent to ComplEx~\cite{HayashiS17}. HolE uses the scoring function $
s(i,k,j) = \vr_k^T (\ve_i \star \ve_j)$, where $\star$ refers to the \emph{circular
  correlation} between $\ve_i$ and $\ve_j\in\mathbb{R}^{d_e}$ (i.e., $ (\ve_i \star
\ve_j)_k=\sum_{t=1}^{d_r} e_{it} e_{j((k+t-2 \mod d_r) + 1)}$). The idea of using
circular correlation relates to associative memory \cite{NickelRP16}. HolE uses
as $\mM^{\text{HolE}}_k$ the circumvent matrix resulting from $\vr_k$. In the
remainder of this paper, we use the formulation using $\mM^{\text{ComplEx}}_k$ 
given above.

\paragraph{\textbf{Analogy}~\cite{LiuWY17}.} Analogy uses block-diagonal mixing
matrices $\mM_k^{\text{Analogy}}$, where each block is either (1) a real scalar $x$
or (2) a $2\times2$ matrix of form $\begin{pmatrix}
  x & -y \\
  y & x
\end{pmatrix}$, where $x,y\in\bR$ refer to entries of $\vr_k$ and each entry of
$\vr_k$ appears in exactly one block. We have $d_r=d_e$. Analogy aims to
capture commutative relational structure: the constraint ensures that
$\mM_{k_1}^{\text{Analogy}}\mM_{k_2}^{\text{Analogy}}=\mM_{k_2}^{\text{Analogy}}\mM_{k_1}^{\text{Analogy}}$ for all $k_1,k_2\in\cR$.
Both DistMult (only first case allowed) and ComplEx (only second case allowed)
are special cases of Analogy.

% \section{Related Work}

% \paragraph{\textbf{Training Trick}}

%%% Local Variables:
%%% mode: latex
%%% TeX-master: "main"
%%% End:

%!TEX root = main.tex
\section{The Relational Tucker3 Decomposition}\label{sec:rtucker}

In this section, we introduce the Relational Tucker3 (RT) decomposition, which
decomposes the KG into entity embeddings, relation embeddings, and a core
tensor. We show that each of the existing bilinear models can be viewed (1) as
an unconstrained RT decomposition with a fixed (sparse) core tensor or (2) as a
constrained RT decomposition with fixed relation embeddings. In contrast to
bilinear models, the RT decomposition allows parameters sharing across different
relations, and it decouples the entity and relation embedding sizes. Our
experimental study (\ref{sec:experiments}) suggests that both properties can be beneficial.
We also introduce a sparse variant of RT called SRT to approach the question of
whether we can learn sparsity patterns of the core tensor from the data.

% We start with the relational Tucker3 decomposition and show that bilinear models are special instances of RTD via the constrained core tensor view and the fixed core tensor view. 
% % This interpretation naturally offers benefits of decoupling entity embedding size and relations embedding size. 
% After that we introduce the sparse relational Tucker3 decomposition.

In what follows, we make use of the tensor representation of KGs. In particular,
we represent knowledge graph $\cK$ via a binary tensor $\tX\in\{0,1\}^{N\times N\times K}$,
where $x_{ijk}=1$ if and only if $(i,k,j)\in\cK$. Note that if $x_{ijk}=0$, we
assume that the truth value of triple $(i,k,j)$ is missing instead of false.
% By convention, vectors $\ve_i$ refer to rows of matrix $\mE$ (as
% a column vector) and scalars $e_{ij}$ to individual entries. 
The \emph{scoring tensor} of a particular embedding model $m$ of $\cK$ is the
tensor $\tS\in\bR^{N\times N\times K}$ of all predicted scores, i.e., with $s_{ijk} =
s^m(i,k,j)$. We use $\mA_k$ to refer to the $k$-th frontal slice of a 3-way
tensor $\tA$. Note that $\mX_k$ contains the data for relation $k$, and that
\emph{scoring matrix} $\mS_k$ contains the respective predicted scores.
Generally, embedding models aim to construct scoring tensors $\tS$ that suitably
approximate $\tX$~\cite{Nickel0TG16}.

\subsection{Definition}
% \begin{figure}
% \centering
% \includegraphics[scale=0.4]{figures/Tucker.pdf}
% \label{fig:RTD}
% \caption{Relational Tucker3 decomposition}
% \end{figure}
% We first recall some basic concepts from tensor algebra~\cite{Tensor}. 
 % \textit{Tensors} are multidimensional arrays.
% \textit{Fibers} are the higher-order analogue of matrix rows and columns. A \textit{mode-$k$ fiber} is defined by fixing every index but the $k$-th. 
% For instance, a row vector of a matrix is a mode-$1$ fiber of a $2$-way tensor and a column vector of a matrix is a mode-$2$ fiber of a $2$-way tensor.
% Mode-$3$ product performs a linear transformation on the mode-$3$ fibers of a tensor:
% \begin{definition}[Mode-$k$ Product]
% \end{definition}
% For the interest of this work, we consider $k\in\{1,2,3\}$ for 3-way tensors.
 % A three-way tensor decomposition models the weighted composition of  three-way interactions. 

We start with the classicial Tucker3 decomposition~\cite{Tucker1966}, focusing
on 3-way tensors throughout. The Tucker3 decomposition decomposes a given data
tensor into three factor matrices (one per mode) and a core tensor, which stores
the weights of the three-way interactions. The decomposition can be can be
viewed as a form of higher-order PCA~\cite{Tensor}.
% \begin{definition}[Tucker3 Decomposition]
In more detail, given a tensor $\tD \in\mathbb{R}^{I\times J\times K}$ and sufficiently
large parameters $d_a,d_b,d_c\in\bN$, the Tucker3 decomposition factorizes $\tD$
into factor matrices $\mA\in \mathbb{R}^{I\times d_a}$, $\mB\in \mathbb{R}^{J\times d_b}$,
$\mC \in \mathbb{R}^{K\times d_c}$, and core tensor $\tH\in\mathbb{R}^{d_a\times d_b\times d_c}$
such that
\[
d_{ijk}  = \va_i^T[\tH\times_3\vc_k]\vb_j,
 % \quad \text{equivalently}, \nonumber \\
% s_{ijk} & = & \sum\limits_{p=1}^{P}\sum\limits_{q=1}^{Q}\sum\limits_{r=1}^{R} h_{pqr}a_{ip}b_{jq}c_{kr} \nonumber.
\]
where $\tH\times_3\vc_k \in \bR^{d_a\times d_b}$ refers to the mode-3 tensor product defined
as
\[
  \tH\times_3\vc_k = \sum_{l=1}^{d_c} c_{kl}\mH_l,
 % [\tH \times_3 \vc]_{i,j} = \sum\limits_{k=1}^{K} h_{i, j, k} c_{k} \quad equiv., \quad [\tH \times_3 \vc]
 % = \sum\limits_{k=1}^{K} \mH_k c_{k}. \nonumber
  % \ \text{for}\  j\in [J'_3]
\]
i.e., a linear combination of the frontal slices of $\tH$. If $d_a, d_b, d_c$
are smaller than $I,J,K$, core tensor $\mathbfcal{H}$ can be interpreted as a
compressed version of $\mathbfcal{D}$. It is well-known that the CP
decomposition~\cite{Tensor}
% (see Sec.~\ref{sec:bilinear}) 
corresponds to the special case
where $d_a=d_b=d_c$ and $\mH$ is fixed to the $d_a\times d_a\times d_a$ tensor with
$h_{ijk}=1$ iff $i=j=k$, else 0. The RT decomposition, which we introduce next,
allows us to view existing bilinear models as decompositions with a fixed core
tensor as well.

In particular, in KG embedding models, we associate each entity with a single
embedding, which we use to represent the entity in both subject and object
position. The relational Tucker3 (RT) decomposition applies this approach to the
Tucker3 decomposition by enforcing $\mA=\mB$. In particular, given embedding
sizes $d_e$ and $d_r$, the RT decomposition is parameterized by an \emph{entity
  embedding matrix} $\mE\in\bR^{N\times d_e}$, a \emph{relation embedding matrix}
$\mR\in\bR^{K\times d_r}$, and a core tensor $\tG\in\bR^{d_e\times d_e\times d_r}$. As in the
standard Tucker3 decomposition, RT composes mixing matrices from the frontal
slices of the core tensor, i.e.,
\[
\mM_k^{\text{RT}} = \sum_{l=1}^{d_r} r_{kl}\mG_l.
\]
The scoring tensor has entries
\begin{align}
  \label{eq:relationaltucker}
  s_{ijk} &= \ve_i^T(\tG\times_3 \vr_k)\ve_j = \ve_i^T\mM_k^{\text{RT}}\ve_j.
\end{align}
Note that the mixing matrices for different relations share parameters through
the frontal slices of the core tensor.

The RT decomposition can represent any given tensor, i.e., the restriction on a
single embedding per entity does not limit expressiveness. To see this, suppose
that we are given a Tucker3 decomposition $\mA,\mB,\mC,\tH$ of some tensor. Now
consider the RT decomposition given by
\begin{align*}
\mE=
  \begin{pmatrix}
   \mA & \mB 
 \end{pmatrix},\
         \mR=\mC,\ \text{and}\
         \mG_{k}=\begin{pmatrix}
           \mathbf{0}_{d_a\times d_a} & \mH_k\\
           \mathbf{0}_{d_b\times d_a}  & \mathbf{0}_{d_b\times d_b}
         \end{pmatrix}.
\end{align*}
We can verify that both decompositions produce the same tensor. Note that we
used a similar construction in Sec.~\ref{sec:bilinear} to represent the CP
decomposition as a bilinear model.

\subsection{The Fixed Core Tensor View}

\begin{figure}
  \centering
  \scalebox{0.7}{
  \begin{tabular}{cccc}
    $ \begin{pmatrix} 1 & 0 \\ 0 & 0 \end{pmatrix}$
    & $ \begin{pmatrix} 0 & 1 \\ 0 & 0 \end{pmatrix}$ 
    & $ \begin{pmatrix} 0 & 0 \\ 1 & 0 \end{pmatrix}$
    & $ \begin{pmatrix} 0 & 0 \\ 0 & 1 \end{pmatrix}$
    \\
    $\mG_1$
    & $\mG_2$
    & $\mG_3$
    & $\mG_4$
  \end{tabular}}
  \caption{Fixed core tensor view of RESCAL ($d_e=2$)}
  \label{fig:rescal}
\end{figure}

\begin{figure}
  \centering
  \scalebox{0.7}{
  \begin{tabular}{cccc}
    $ \begin{pmatrix} 1 & 0 & 0 & 0 \\ 0 & 0 & 0 & 0 \\ 0 & 0 & 1 & 0 \\ 0 & 0 & 0 & 0\end{pmatrix}$
    & $ \begin{pmatrix} 0 & 0 & 0 & 0 \\ 0 & 1 & 0  & 0 \\ 0 & 0 & 0 & 0 \\ 0 & 0 & 0 & 1\end{pmatrix}$
    & $ \begin{pmatrix} 0 & 0 & 1 & 0 \\ 0 & 0 & 0 & 0 \\ -1 & 0 & 0 & 0 \\ 0 & 0 & 0 & 0\end{pmatrix}$
    & $ \begin{pmatrix} 0 & 0 & 0 & 0 \\ 0 & 0 & 0 & 1 \\ 0 & 0 & 0 & 0 \\ 0 & -1 & 0 & 0\end{pmatrix}$
    \\
    $\mG_1$
                       & $\mG_2$
                            & $\mG_3$
                            & $\mG_4$
  \end{tabular}}
  \caption{Fixed core tensor view of ComplEx ($d_e=4$) 
  % \todo{If space left, make larger and two rows}
  }
  \label{fig:complexfrontal}
\end{figure}

The RT decomposition gives rise to a new interpretation of the bilinear models
of Sec.~\ref{sec:bilinear}: We can view them as RT decompositions with a fixed
core tensor and unconstrained entity and relation embedding matrices.

Intuitively, in this fixed core tensor view, the relation embedding $\vr_k$
carries the relation-specific parameters as before, and the core tensor describes
where to ``place'' these parameters in the mixing matrix. For example, in
RESCAL, we have $d_r=d_e^2$, and each entry of $\vr_k$ is placed at a separate
position in the mixing matrix. We can express this placement via a fixed core
tensor $\tG^{\text{RESCAL}}\in\mathbb{R}^{d_e\times d_e\times d_r}$ with
\[
g_{ijk}^{\text{RESCAL}} = \begin{cases}
      1 & \text{if $i=\floor{\frac{k-1}{d_e}}+1$ and $j=(k-1\bmod d_e)+1$} \\
      0 & \text{otherwise}
    \end{cases}.
\]
% \todo{fixed; please double check; needs to match fig1!} 
The frontal slices of
$\tG^{\text{RESCAL}}$ for $d_e=2$ (and thus $d_r=4$) are shown in
Fig.~\ref{fig:rescal}. As another example, we can use a similar construction for
ComplEx, where we use $\tG^{\text{ComplEx}}\in \mathbb{N}^{d_e\times d_e\times d_e}$ with
\[
g_{ijk}^{\text{ComplEx}} = \begin{cases}
      1 & \text{if $1\leq k \leq \frac{d_e}{2}$ and $i=j=k$,}\\
        & \text{or $1\leq k \leq \frac{d_e}{2}$ and $i=j=k+\frac{d_e}{2}$,}\\
        & \text{or $\frac{d_e}{2}+1 \leq k \leq d_e$ and $i=k-\frac{d_e}{2}$ and $j=k$}\\
      -1 & \text{if $\frac{d_e}{2}+1 \leq k \leq d_e$, $i=k$, and $j=k-\frac{d_e}{2}$} \\
      0 & \text{otherwise}.
    \end{cases}
\]
The corresponding frontal slices for $d_e=4$ are illustrated in Fig.~\ref{fig:complexfrontal}. % In terms of representation, we can treat Tucker3 decomposition as a bilinear model.
% \begin{figure*}%
% \centering
% \subfigure[Relational Tucker3 decomposition]{\includegraphics[scale=0.3]{figures/Tucker.pdf}}
% \label{fig:RTD}
% \qquad
% \subfigure[ComplEx core tensor in the constrained core tensor view ($d=d_e$)]{\includegraphics[scale=0.2]{figures/sparsecore3.pdf}}
% \label{fig:complexcoretensor}
% \caption{RTD and ComplEx core tensor} 
% \end{figure*}

% \begin{figure}
% \centering
% \includegraphics[scale=0.23]{figures/sparsecore3.pdf}
% \caption{ComplEx core tensor in the constrained core tensor view ($d=d_e$)} 
% \label{fig:complexcoretensor}
% \end{figure}

If we express prior bilinear models via the fixed core tensor viewpoint, we
obtain extremely sparse core tensors. The sparsity pattern of the core tensor is
fixed though, and differs across bilinear models. A natural question is whether
or not we we can learn the sparsity pattern from the data instead of fixing it
a priori, and whether and when such an approach is beneficial. We empirically
approach this question in our experimental study in Sec~\ref{sec:experiments}.

% \subsection{Relational Tucker Decomposition}
\subsection{The Constrained Core Tensor View}\label{sec:bitucker}

An alternate viewpoint of existing bilinear models is in terms of a constrained
core tensor. In this viewpoint, the relation embedding matrix is fixed to the
$K\times K$ identity matrix, the entity embedding matrix is unconstrained. We have
\[
s_{ijk}  = \ve_i^T[\tG\times_3\vi_k]\ve_j = \ve_i^T\mG_k\ve_j.
\]
The core tensor thus contains the mixing matrices directly, i.e., $\mM_k=\mG_k$.
The various bilinear models can be expressed by constraining the frontal slices
of the core tensor appropriately (as in Sec~\ref{sec:bilinear}).

\subsection{Discussion}

One of the main difference of both viewpoints is that in the fixed core tensor
viewpoint, $d_r$ is determined by $d_e$ (e.g., $d_r=d_e$ for ComplEx) and
generally independent of the number of relations $K$. If $d_r<K$, we perform
compression along the third mode (corresponding to relations). In contrast, in
the constrained core tensor viewpoint, we have $d_r=K$ so that no compression of
the third mode is performed.

In a general RT decomposition, there is no a priori coupling between the entity
embedding size $d_e$ and the relation embedding size $d_r$, which allows us to
choose $d_r$ freely. Moreover, since the core tensor is shared across relations,
different mixing matrices depend on shared parameters. This property can be
beneficial if dependencies exists between relations. To illustrate this point,
assume a relational dataset containing the three relations
% \textit{based\_in, located\_in} and \textit{headquarter}, e.g., \textit{(Siemens, based in, Munich), (ABB, located\_in, Zurich)} and \textit{(Google, headquarter, California)}. 
\textit{parent} ($p$), \textit{mother} ($m$), and \textit{father} ($f$). Since
generally $\textit{parent}(i,j) \iff \textit{mother}(i,j) \vee \textit{father(i,j)}$,
the relations are highly dependent. Suppose for simplicity that there exists
mixing matrices $\mM_m$ and $\mM_f$ that perfectly reconstruct the data in that
$\ve_i^T\mM_m\ve_j=x_{ijm}$ (likewise $f$). If we set $\mM_p=\mM_m+\mM_f$, then
$x_{ijp}\ge 0 \iff \ve_i^T\mM_p\ve_j\ge 0$, i.e., we can reconstruct the
\textit{parent} relation without additional parameters. We can express this with
an RT decomposition with $d_r=2<3=K$, where $\vr_m= \begin{pmatrix} 1 &
  0\end{pmatrix} $, $\vr_f= \begin{pmatrix} 0 & 1\end{pmatrix} $,
$\vr_p= \begin{pmatrix} 1 & 1\end{pmatrix} $, $\mG_1=\mM_m$, and $\mG_2=\mM_f$.
% In addition, assume the dependency
% % \begin{eqnarray}
% % headquarter(i,j) \Rightarrow located\_in(i,j) \nonumber \\
% % based\_in(i,j) \Rightarrow located\_in(i,j) \nonumber \\
% %  located\_in(i,j) \Rightarrow headquarter(i,j)\lor based\_in(i,j) \nonumber \\
% % headquarter(i,j)\land based\_in(i,j) = \emptyset \nonumber
% % \end{eqnarray}
% \begin{eqnarray}
% parent(i,j) \Leftrightarrow mother(i,j) \  \text{or} \  father(i,j). \nonumber 
% % based\_in(i,j) \Rightarrow located\_in(i,j) \nonumber \\
%  % located\_in(i,j) \Rightarrow headquarter(i,j)\lor based\_in(i,j) \nonumber \\
% % headquarter(i,j)\land based\_in(i,j) = \emptyset \nonumber
% \end{eqnarray}
% holds in the dataset.
% We can model such a dataset with an RT whose core tensor $\tG$ containing $2$ frontal slices instead of $3$ (one for each relation as in existing bilinear models).
% i.e., we do not need to learn embeddings for every single relation to represent the corresponding knowledge graph $\tK$. 
% To see this, suppose that relation \textit{mother} and \textit{father} are modeled well with $\tG, \vr_{father}, \vr_{mother}$, i.e., the scoring matrices are good approximations for the corresponding data matrices ($\mS_{father}\approx \mX_{father}$ and $\mS_{mother}\approx \mX_{mother}$). 
% Setting $\vr_{parents}=\vr_{mother}+\vr_{father}$,  we have \[\mS_{parent}=\mS_{father}+\mS_{mother}\approx \mX_{father}+ \mX_{mother}=\mX_{parent}.\]
By choosing $d_r<K$, we thus compress the frontal slices and force the model to
discover commonalities between the various relations.

Since the mixing matrix of each relation is determined by both the relation
embeddings $\mR$ and the core tensor $\tG$, an RT decomposition can have many
more parameters per relation than $d_r$. To effectively compare various models,
we define the \textit{effective relation embedding size} $d_r^*$ of a given RT
decomposition as the average number of parameters per relation. More precisely,
we set
\begin{equation}\label{eq:effective}
d_r^*=\frac{\operatorname{nnfp}(\tG)+\operatorname{nnfp}(\mR)}{K}, \nonumber
\end{equation}
where $\operatorname{nnfp}$ refers to the number of \textit{non-zero free
  parameters} in its argument. This definition ensures that the effective
relation embedding size of a bilinear model is identical under both the fixed
and the constrained core tensor interpretation (even though $d_r$ differs).
Consider, for example, a ComplEx model. Under the fixed core tensor viewpoint,
we have $\operatorname{nnfp}(\tG)=0$ and $\operatorname{nnfp}(\mR)=Kd_e$ so that
$d_r^*=d_e$. In the constrained core tensor viewpoint, we have
$\operatorname{nnfp}(\mR)=0$ and $\operatorname{nnfp}(\tG)=Kd_e$ so that
$d_r^*=d_e$ as well (although $d_r=K$). For RESCAL, we have $d_r^*=d_e^2$.
% More prec for the relations, i.e., then $d_r^*$ of RESCAL is $d_e^2$ since
% $\operatorname{nnfp}(\tG)=Kd_e^2$ and $\operatorname{nnfp}(\mR)=0$, while it is
% $d_e$ for ComplEx, DistMult, HolE and Analogy since
% $\operatorname{nnfp}(\tG)=Kd_e$ and $\operatorname{nnfp}(\mR)=0$.
% In this regard, existing bilinear models, despite their differences, only
% considered two cases of effective relation embedding sizes.
For a fixed entity embedding size $d_e$, it is plausible that the suitable
choice of $d_r^*$ is data-dependent. In the RT decomposition, we can control
$d_r^*$ via $d_r$, and thus also decouple the entity embedding size from the
effective relation embedding sizes.
% If we relax $\mR$, adjusting $d_r$ allows us to control $d_r^*$ and offers a
% flexible way to decouple $d_e$ and $d_r^*$.
The \textbf{effective number of parameters} of an RT model is given by
\[
\operatorname{nnfp}(\tG)+\operatorname{nnfp}(\mR)+\operatorname{nnfp}(\mE).
\]

\subsection{Sparse Relational Tucker Decomposition}
% \paragraph{Motivation for learning sparse core tensor} 
\label{sec:autosparse}

In bilinear models such as ComplEx, DistMult, Analogy, or the CP decomposition,
the core tensor is extremely sparse under both interpretations. In a general RT
decomposition, this may not be the case and, in fact, the core tensor can become
excessively large if it is dense and $d_e$ and $d_r$ are large (it has $d_e^2d_r$
entries). On the other hand, the RT decomposition allows to share parameters
across relations so that we may use significantly smaller values of $d_r$ to
obtain suitable representations. In our experimental study, we found that this
was indeed the case in certain settings.

To explore the question of whether and when we can learn sparsity patterns from
the data instead of fixing them upfront, we make use of a sparse RT (SRT)
decomposition, i.e., an RT decomposition with a sparse core tensor.
Let $\vtheta=\{\mE, \mR, \tG\}$ be the parameter set and 
% \[\cL:\mathbb{B}^{N\times N\times K} \times \mathbb{R}^{N\times N\times K} \rightarrow \mathbb{R}\] 
$\cL$ be a loss function. In SRT, we add an additional $l_0$ regularization term
on the core tensor and optimize
\begin{align}\label{eq:sparse}
% \textcolor{red}{\cR(\vtheta)=\sum\limits_{i,j,k}\cL(X_{ijk},\ve^T_i[\tH\times_3\vr_k]\ve_j)+\lambda\|\tH\|_0}.\\
\cR(\vtheta)=\cL(X,\vtheta)+\lambda\|\tG\|_0.
\end{align}
where $\lambda$ is an regularization hyper-parameter. Solving Eq.~\eqref{eq:sparse}
exactly is NP-hard. In practice, to obtain an approximate solution, we apply the
\textit{hard concrete} approximation~\cite{Concrete,L0norm}, which has shown
good results on sparsifying neural networks.\footnote{Other sparsification
  techniques can be applied as well, of course, but we found this one to work
  well in practice.} This approach also allows us to maintain a sparse model
during the training process. 
% \todo{Really?!}
% Our $l_0$ formulation corresponds to the constrained core tensor view as entries of $\tG$ does not have to be binary. 
In contrast to prior models, the frontal slices of the learned core tensor can
have different sparsity patterns, capturing the distinct shared components. 

\section{Experiments}\label{sec:experiments}

We conducted an experimental study on common benchmark datasets to gain insight
into the RT decompositon and its comparison to the state-of-the-art model
ComplEx. Our main goal was empirically study whether and when we can learn
sparsity patterns from the data, and whether sparsity is necessary. We compared
dense RT (DRT) decompositions, sparse RT (SRT) decompositions, and ComplEx
w.r.t.~(1) best prediction performance overall, (2) the relationship between
entity embedding size and prediction performance, and (3) the relationship
between model size (in terms of effective number of parameters) and prediction
performance.

We found that an SRT \emph{can} perform similar or better than ComplEx,
indicating that it is sometimes possible and even beneficial to learn the
sparsity pattern. Likewise, we observed that a DRT \emph{can} outperform both
SRT and ComplEx with a similar effective number of parameters and with only a
fraction of the entity embedding size. This is not always the case, though: the
best model generally depends on the dataset and model size requirements.

% By analyzing the learned sparsity patterns of a SRT model, we found
% that the diagonal entries in the resulting mixing matrices are not pronounced in
% many frequent relations, in which case SRT performed better than ComplEx.

% Gor a given effective number of parameters,   % and this insight is supported by a simple yet competitive model DCP. 
 % in terms of both performance and parameter efficiency; 
% \textcolor{red}{add DCP summary}
% We implemented SRT and ComplEx using PyTorch and conduct experiments on benchmark datasets to investigate the  proposed questions.  The code for the experiments will be made available. 
\subsection{Experiment Setup}

\paragraph{Data and evaluation.}
We followed the widely adopted entity ranking evaluation
protocol~\cite{BordesUGWY13} on two benchmark datasets: FB15K-237 and
WN18RR~\cite{DBLP:conf/emnlp/ToutanovaCPPCG15,dettmers2018conve}. The datasets
are subsets of the larger WN18 and FB15K datasets, which are derived from
WordNet and Freebase respectively~\cite{BordesUGWY13}. Since WN18 and FB15K can
be modeled well by simple rules~\cite{dettmers2018conve,meilicke2018fine},
FB15K-237 and WN18RR were constructed to be more challenging. See
Tab.~\ref{tab:data} for key statistics. In \emph{entity ranking}, we rank
entities for test queries of the form $k(?,e)$ or $k(e,?)$. We report the
\emph{mean reciprocal rank (MRR)} and \emph{HITS@k} in the \emph{filtered}
setting, in which predictions that correspond to tuples in the training or
validation datasets are discarded (so that only new predictions are evaluated).
%The key data statistics are in the appendix.

% We evaluated model performance via entity ranking on the test data. 

% Let $\mathcal{T}^{train}, \mathcal{T}^{val}, \mathcal{T}^{test}$ be the training, validation, and test
% data, resp. Then FB15K-237 is obtained from FB15k by removing inverse relations and
% by ensuring that whenever $(i,k,j)\in \mathcal{T}^{test} \cup \mathcal{T}^{val}$,
% $(i,k',j)\notin \mathcal{T}^{train}$ for all $k'\neq k$. WN18RR is constructed from 
% WN18 by following the same procedure.
%  Key dataset statistics are summarized in 
% Table~\ref{tab:data}.
 \begin{table}
   \centering
   \caption{Dataset statistics}
   \label{tab:data}
   \begin{tabular}{l@{\hspace{.7em}}r@{\hspace{.7em}}r@{\hspace{.7em}}r@{\hspace{.7em}}r@{\hspace{.7em}}r}
     \hline\hline
     Dataset & \# Ent & \# Rel         & \# Train  &        \# Valid    &    \# Test          \\ \hline
     FB15K-237   & \numprint{14505} & \numprint{237} &  \numprint{272115}  &  \numprint{17535}  &  \numprint{20466} \\ 
       WN18RR    & \numprint{40559} & \numprint{11} &  \numprint{86835}  &  \numprint{2824}   &  \numprint{2924}  \\
     \hline\hline
   \end{tabular}
 \end{table}
\begin{table*}[ht]
  \centering
  \caption{Best entity ranking results in terms of MRR for $d_e\le200$. Note that
    for WN18RR, the effective number of parameters does not change significantly
    with increasing effective relation size because it has only $11$ relations.
  }
  \label{tab:selfcomparison237}
    \begin{tabular}{lrrrrrrrr}
      \hline\hline
                        & Entity & Relation  & Effective relation & Effective number   & MRR  & \multicolumn{3}{c}{HITS}  \\ 
    Model                 & size & size $d_r$ &  size $d_r^*$ & of parameters &  & @1 & @3 & @10   \\ \hline
    & \multicolumn{8}{c}{FB15K-237}\\ \hline 
    ComplEx & 200 & -   & 200    & 3,047K & 27.2 & 17.6 & 30.4 & 46.7  \\  
    DRT      & 100 & 237 & 10,237 & 3,926K & 28.4 & 19.2 & 31.2 & 47.2 \\
    SRT     & 200 & 200 & 16,401 &  6,887K   &  28.5     & 19.2    &  31.7      &  47.3   \\     \hline
    & \multicolumn{8}{c}{WN18RR}\\ \hline 
%            & entity & relation & effective     & \#parameters  & MRR           & \multicolumn{3}{c}{HITS}  \\ 
%    Model   & size   & size     & relation size &               &               & @1   & @3 & @10   \\ \hline
    ComplEx & 200    & -      & 200           & 8,114K        & 47.0          & 42.0 & 50.0 & 55.4   \\
    DRT      & 200    & 11       & 40,011        & 8,552K        & 41.9          & 40.0 & 42.8 & 45.2  \\  
    SRT     & 200    & 7        & 2,909         & 8,144K        & 42.5          & 40.0 & 43.9 & 46.8  \\
    \hline
\hline
    \end{tabular}
\end{table*}

\paragraph{Models and training.} We implemented DRT, SRT, and ComplEx using
PyTorch.
% ComplEx has been show to be very competitive \citep{dettmers2018conve,Canonical} and yet simple to implement. 
Our ComplEx implementation provides a very strong baseline; it achieves
state-of-the-art results (see Sec.~\ref{sec:compvsprior}) with far smaller
embedding sizes than previously reported. We trained all models with
AdaGrad~\cite{DuchiHS11} using cross-entropy loss with negative sampling. In
each step, we sampled a batch of positive triples at random and obtained
pseudo-negative triples for each positive triple by randomly perturbing the
subject or object of each positive triple. Sampling was done without
replacement. This common approach corresponds to the locally-closed world
assumption in the literature~\cite{Nickel0TG16}. We computed the scores of each
positive triple \textit{(i,k,j)}
% score function
% \[
% s(i,k,j|\bar{\theta})=\ve_i[\tH\times_3\vr_k]\ve_j,
% \] t, a true triple $(i,k,j)$ 
and its associated pseudo-negative triples, applied softmax, and used
cross-entropy loss on the result.
%\begin{align*}
%   & P(i,k,j) = \frac{\exp(s(i,k,j))}{\exp(s(i,k,j))+\sum\limits_{(i',k,j') \in 
%   \mathcal{N}_{ijk}}\exp(s(i',k,j'))}
%\end{align*}
% ($P(j|i,k)$ is computed similarly). 
%In the end, the cross-entropy loss is applied.
%For SRT, let  $\phi_{ijk}$ be the parameters of a hard concrete variable $Z_{ijk}$  and $\Phi$ be the union for all $\phi_{ijk}$'s. We optimize
%\begin{align*}
%      -\mathbb{E}_{q(\tZ|\Phi)}  \bigl[ 
%    \sum_{(i,k,j)\in \cK} \log(P(i,k,j))\bigr]
%        +
%\lambda \sum\limits_{i,j,k}(1-F_{\phi_{ijk}}(0))
%\end{align*}
%as an approximation for Eq. ~\eqref{eq:sparse}, where $F_{\phi_{ijk}}(\cdot)$ is the CDF of a hard concrete variable~\cite{L0norm,Concrete}. 
% The quantity $1-F_{\phi_{ijk}}(0)$ is the relaxed counterpart of the success probability in the Bernoulli distribution. 
% To approximate the $l_0$ regularization for SRT (see Sec. \ref{sec:autosparse}), 
To approximate the $l_0$ regularization for SRT (see Sec. \ref{sec:autosparse}),
we adapted an existing
implementation\footnote{\url{https://github.com/moskomule/l0.pytorch/}} to our
setting.
% \todo{dropped reciprocal: not needed here anymore}
% Recent works use a method which augment the training data and learn two embeddings per relation.
% i.e., reciprocal trick~\cite{Canonical}
% , i.e. reciprocal relations  for details), that 
% empirically improves the results. 
% We did not apply this method for all models throughout the paper (See Sec.~\ref{sec:complexbaseline} for a discussion).

\paragraph{Hyperparameters and model selection.} 

%\citet{L0norm}.   
Previous work~\cite{NickelRP16,TrouillonN17,Canonical} has shown that model
performance is sensitive to loss function and hyperparameters. We consistently
used cross-entropy loss with $24$ pseudo-negative samples ($i,j$ each) for all
models for a fair comparison. The batch size was fixed to $500$ whenever
possible. The only exception was the SRT model with an entity embedding size of
200, where we used a batch size of $256$ due to GPU memory constraints.

To compare ComplEx, DRT, and SRT within similar entity embedding sizes, we fixed
the the entity embedding size $d_e$ and tuned all other hyperparameters using
Bayesian Optimization.\footnote{We use scikit-learn's GaussianProcessRegressor
  (v 0.20.1) with its defaults settings with 20 restarts.} To keep our study
feasible, we only considered RT with entity embedding sizes $d_e\leq200$.
% We took $8$ distinct random samples from the grid of hyper parameter settings and then explore $4$ unseen hyper parameter settings that receive the highest predicted score plus standard deviation.
In total, we evaluated $12$ hyperparameter settings for each entity embedding
size. For all models, we searched over the dropout $\eta$ on the entity embeddings
and mixing matrixes, the learning rate $lr$, and the weight decay $\omega$. For SRT
and RT, we additionally searched over the relation embedding sizes $d_r$ to
study the effect of compression of the third mode. Moreover, for SRT, we
searched over the $l_0$ regularization parameter $\lambda$, which determines the
degree of sparsity. A summary of all hyperparameter ranges and additional training details are given in the supplementary material. 
% \todo{Keep in mind}
% \footnote{For SRT we set $\lambda=0$ for the first $25$ epochs, i.e. we initially do not include the $l_0$ regularization term, as it empirically  improved the results.} (see \textcolor{red}{appendix} for the ranges). 
Model selection was based on MRR on validation data using early stopping (no
improvement for 10 epochs).
% \todo{Why mention and what does it mean?: }Notice that this does not necessarily favor very sparse SRTs.

\subsection{Results}

% \todo{Dropped intro sentence because it's now in intro to entire section. Readd if you want.}

\paragraph{Prediction performance.} Tab.~\ref{tab:selfcomparison237} reports the
best performance (in terms of MRR) obtained by each model for FB15K-237 and
WN18RR. We restrict the comparison to entity embedding sizes $\leq200$, for which
training RT was feasible on our hardware.

On FB15K-237, SRT and DRT showed similar performance, suggesting that sparsity
is not always necessary. Indeed, the core tensor of the best-performing SRT
solution was only 48\% sparse. Moreover, both models were competitive to
ComplEx, suggesting that it is possible to learn suitable models without
imposing a fixed sparsity pattern. The best-performing SRT used $d_r=200<K=237$
and thus performed (some) compression on the relation embeddings.

The results differ significantly on WN18RR. Here, SRT performed slightly better
than DRT, while using a smaller relation embedding size ($d_r=7<11=K$) and used
a much smaller effective relation embedding size. Nevertheless, ComplEx was
clearly the best performing model on this dataset; the fixed sparsity pattern
seems to help. SRT and DRT are close to ComplEx only for HITS@$1$.

Overall, no model consistently outperformed all the others on all datasets.

\begin{figure}[t]%
  \centering \subfigure[Influence of entity embedding
  size]{\includegraphics[scale=0.5]{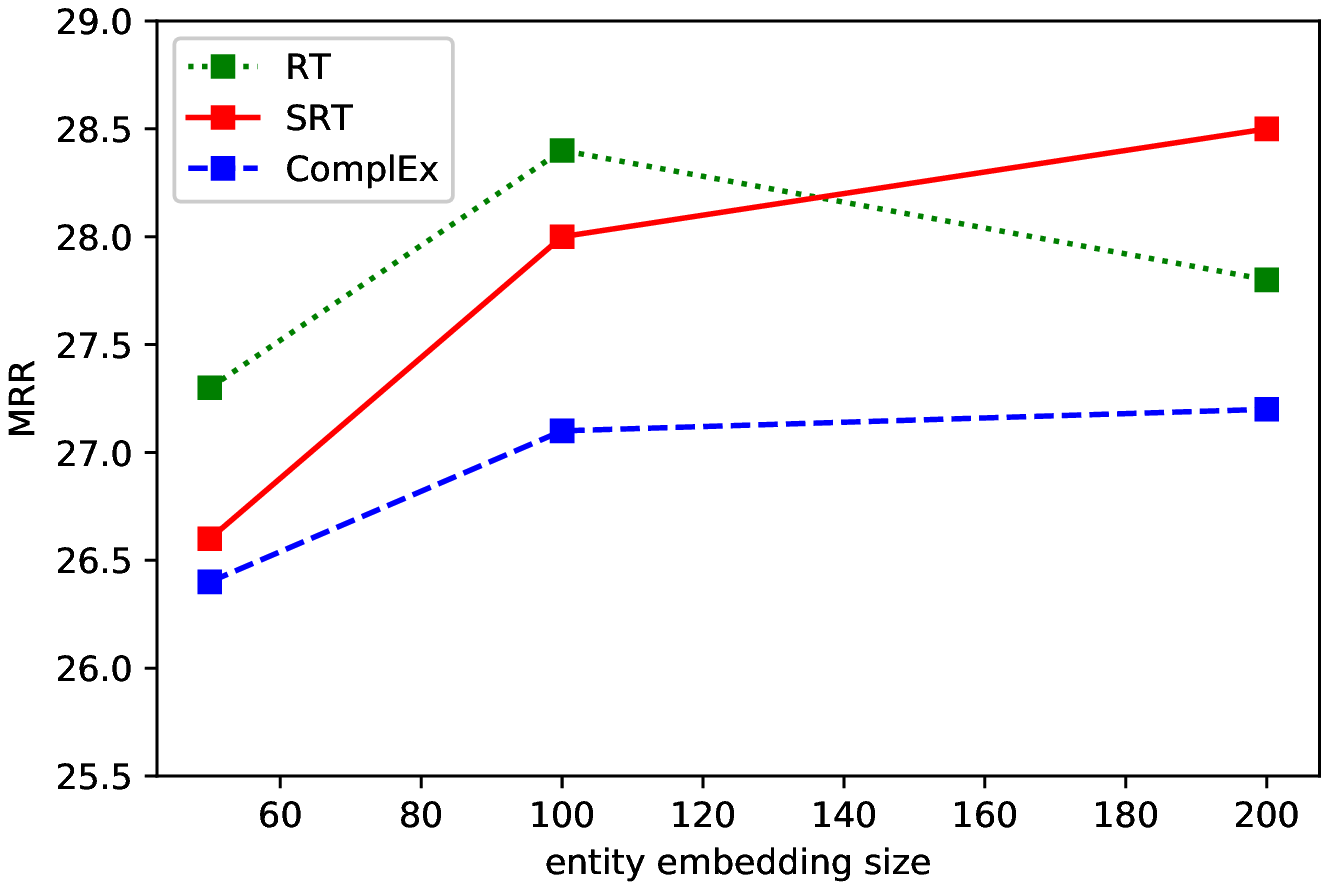}} \qquad
  \subfigure[Influence of model size. For RT, $d_e=32,52,84$ (from left to right).
  For SRT and ComplEx, $d_e=50,100,200$ 
  % \todo{not true, SRT has more points. Add
    % the $d_e$ sizes to the plot in the corresponding
    % colors.} \todo{The x-axis is messed up! write label ``effective number of parameters (millions)'' and drop the repeated $10^6$}
    ]{\includegraphics[scale=0.5]{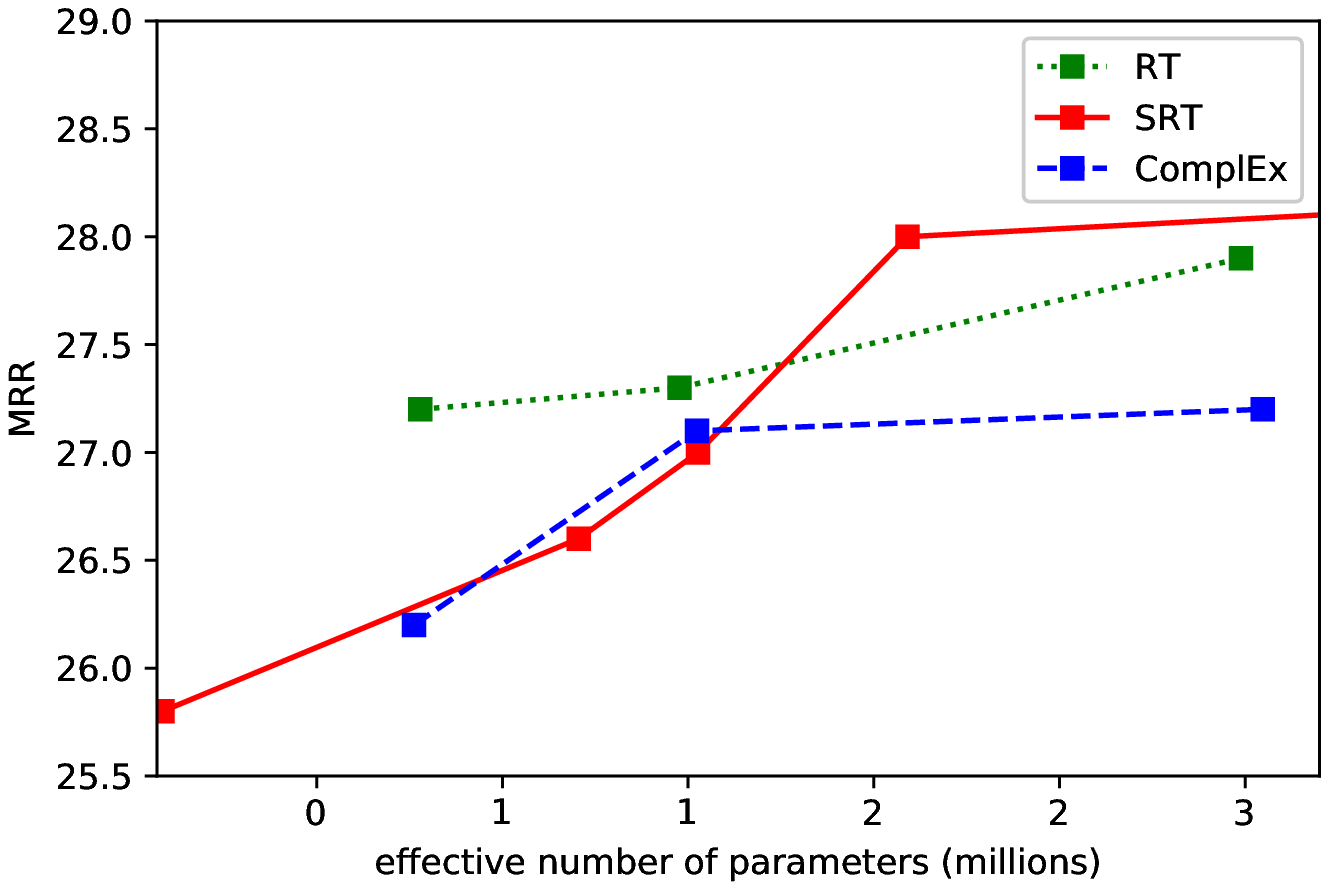}}
\caption{Performance analysis on FB15K-237} 
\label{fig:budgetfb}
\end{figure}

\begin{figure}[t]%
  \centering \subfigure[Influence of entity embedding
  size]{\includegraphics[scale=0.5]{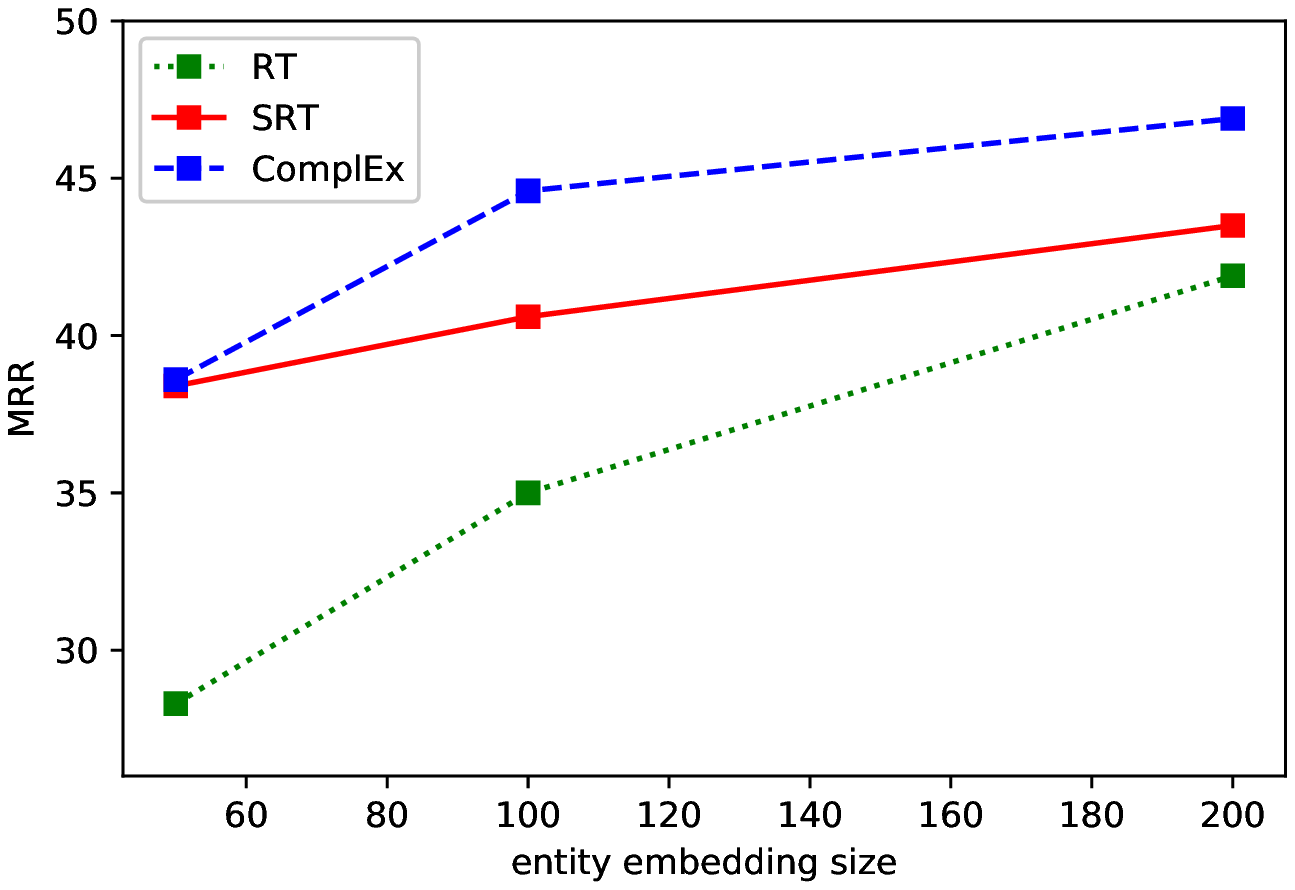}} \qquad
  \subfigure[Influence of model size. For all models, $d_e=50,100,200$ from left to
  right.]{\includegraphics[scale=0.5]{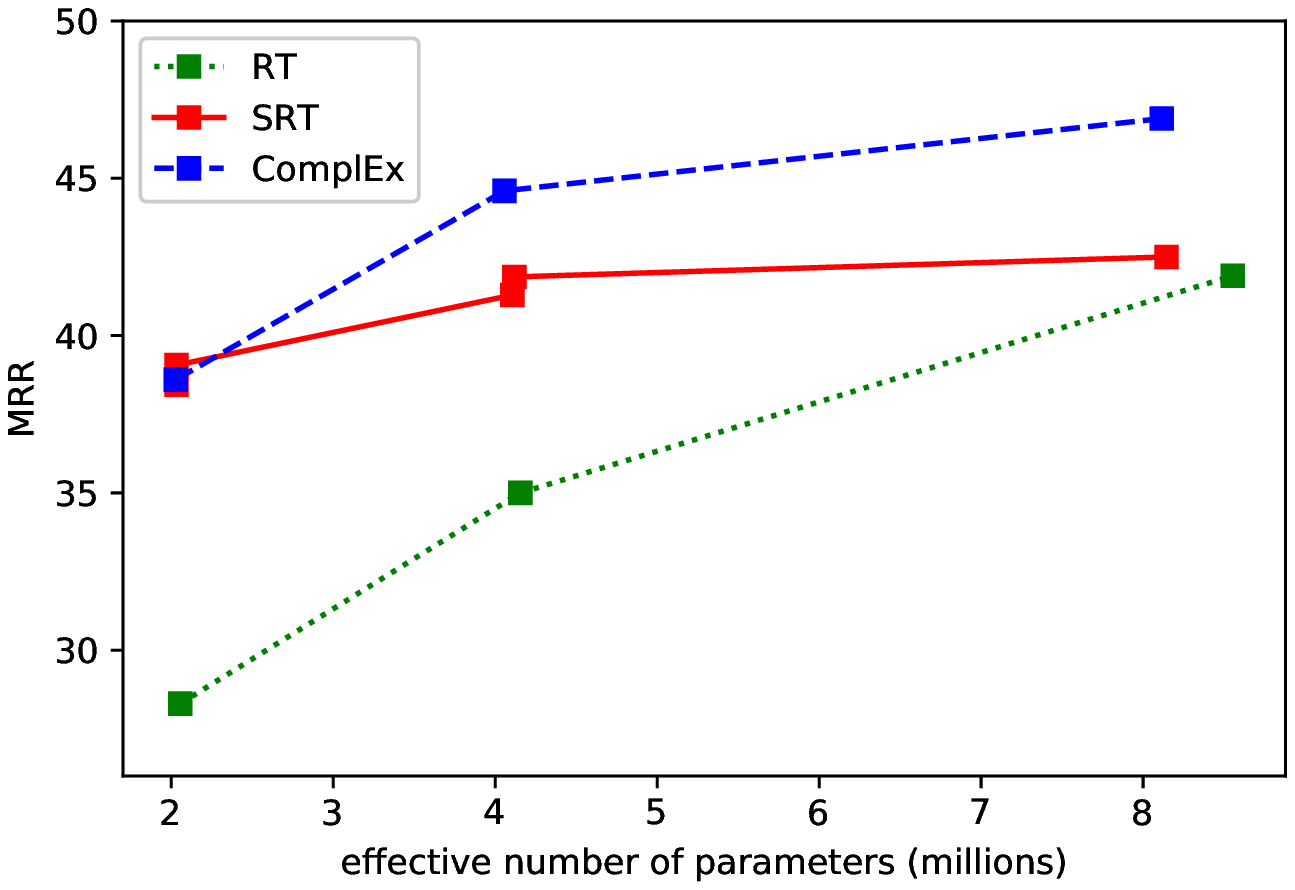}}
\caption{Performance analysis on WN18RR 
% \todo{resolution still somewhat. why not
    % include a PDF?}
}
\label{fig:budgetwn}
\end{figure}

\paragraph{Influence of entity embedding size.}
Figures~\ref{fig:budgetfb}a) and~\ref{fig:budgetwn}a) report model
performance for varying entity embedding sizes. For FB15K-237,
Figure~\ref{fig:budgetfb}(a) shows that both SRT and DRT performed competitive
or better than ComplEx across the entire range of $d_e\in\{50,100,200\}$. Sparsity
constraints were helpful when $d_e\geq 100$: the performance of DRT decreased while
SRT continued to improve on WN18RR.
% , thus sparsity helps when models with equal entity embedding size are compared. 

\paragraph{Influence of effective number of parameters.}
Figures~\ref{fig:budgetfb}b) and~\ref{fig:budgetwn}b) report model performance
for varying model sizes, measured in terms of the effective number of
parameters. The plot is a Skyline plot, i.e., we do not include model fits that
are outperformed by another fit of the same model with smaller effective number
of parameters.

On FB15K-237, DRT consistently outperformed ComplEx, using a smaller entity
embedding size but (consequently) a larger effective relation
size.
% \footnote{Additional hyper-parameters to ensure similar parameter budget by
  % setting $d_r=K$. \todo{?????????}} 
  Sparsity thus does not always help, and
parameter sharing seemed to be beneficial. For small models, SRT performed
similar to ComplEx; for larger models, SRT performed better than both ComplEx
and DRT. This suggests that it is indeed possible learn sparsity patterns from
the data, and that this can be beneficial.
 % \todo{make sure to put these points
  % in the intro} 
  On WN18RR, ComplEx performed best as in our previous
experiments. On both datasets, we found that SRT models with smaller entity
embedding size $d_e$ usually had denser core tensors. We conjecture that this
allows the SRT model to ``assign'' more responsibility to the core tensor with
similar number of parameters.

Overall, 
% \todo{quick summary of everything, e.g., modify the stuff below} 
this
comparison shows decoupling $d_e$ and $d_r$ can lead to competitive models with very small entity embedding size
% \todo{why?}, 
and
that sparsity does not always help when models with equal parameter sizes are
compared. On the other hand, for some datasets the prescribed constraints can be
very effective.

\subsection{Learned Sparsity Patterns}

\begin{figure}[t]%
\centering
\subfigure[win\_same\_award]{\includegraphics[scale=0.5]{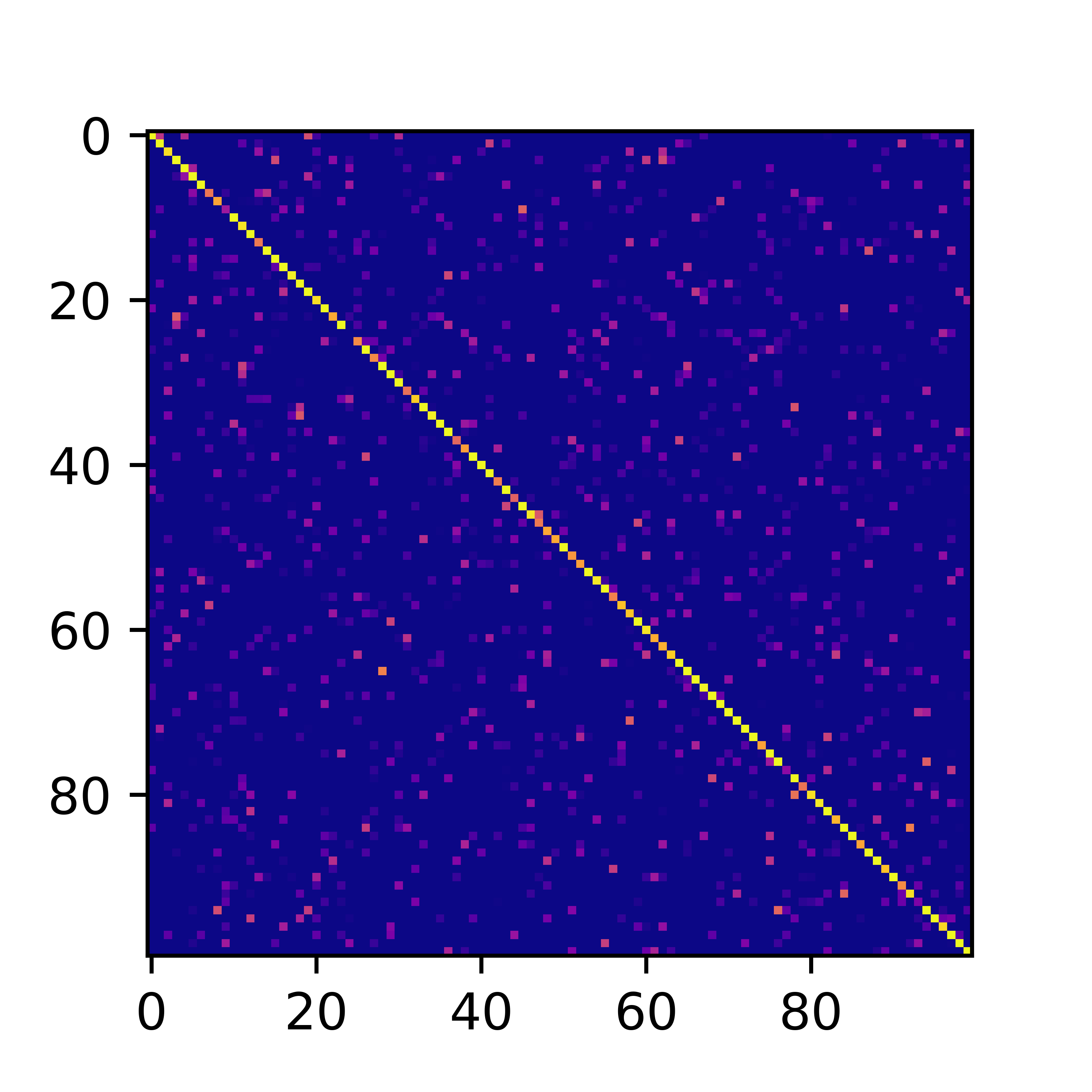}} \quad
\subfigure[location\_contains]{\includegraphics[scale=0.5]{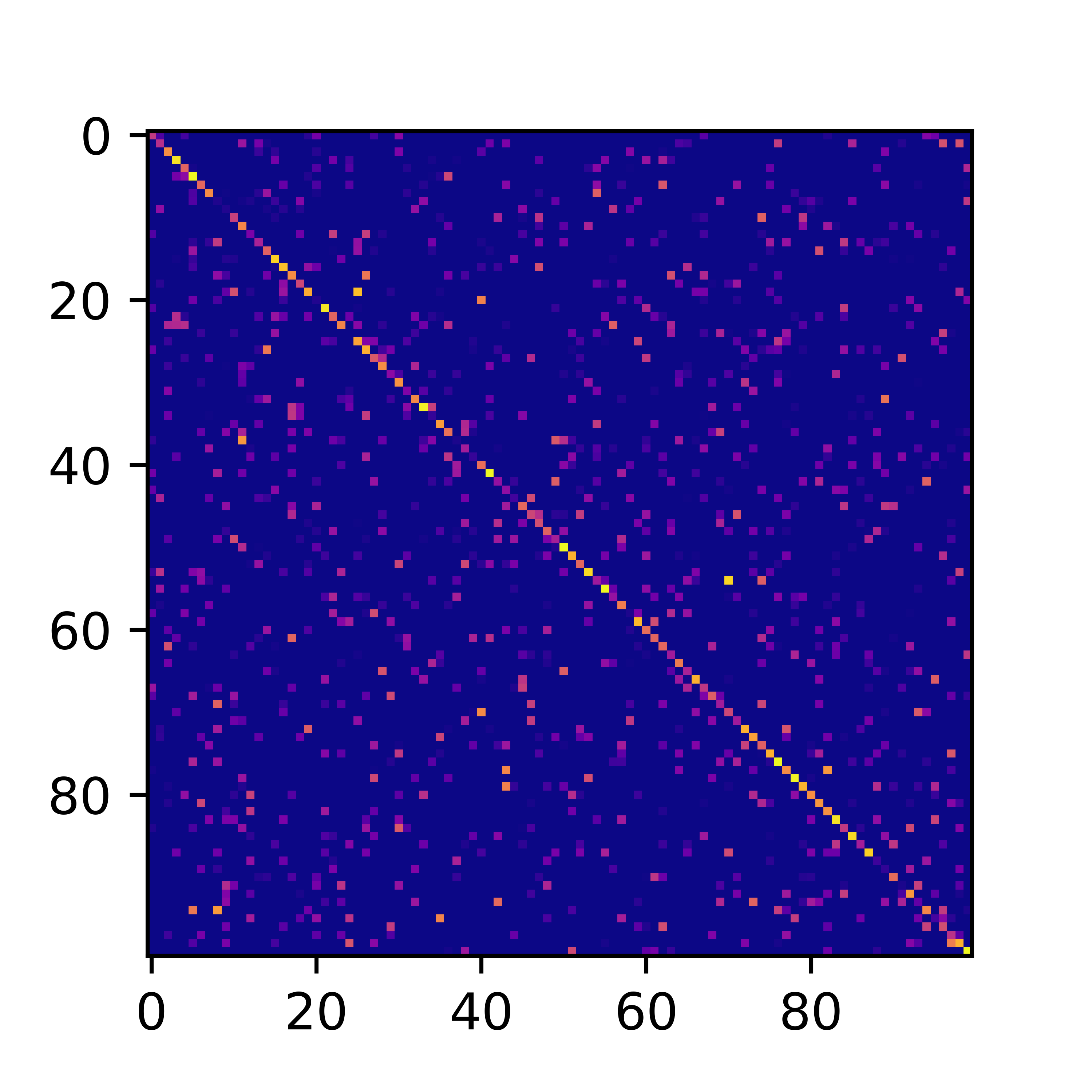}} \quad
\subfigure[type\_of\_union]{\includegraphics[scale=0.5]{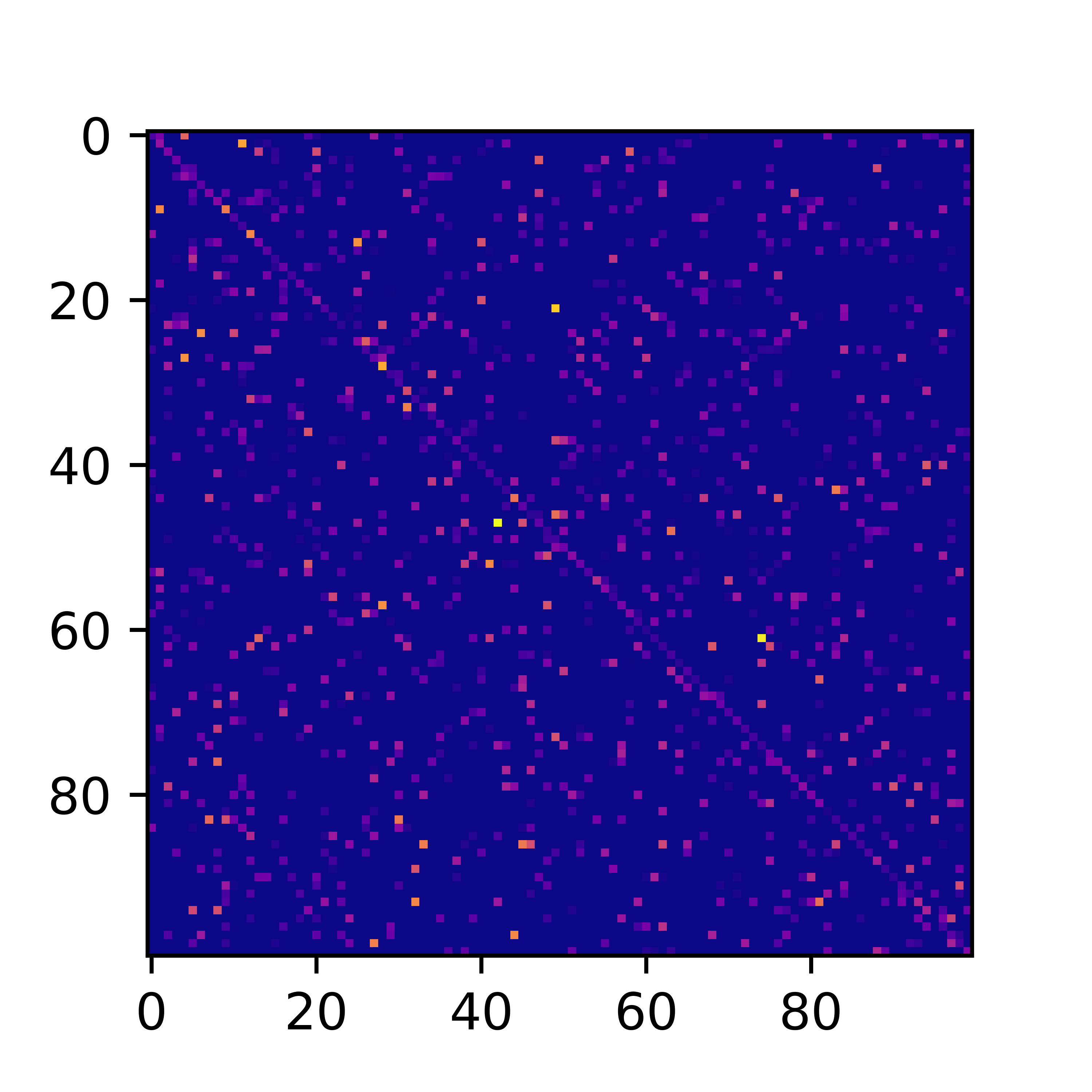}}  \quad
\subfigure[gender]{\includegraphics[scale=0.5]{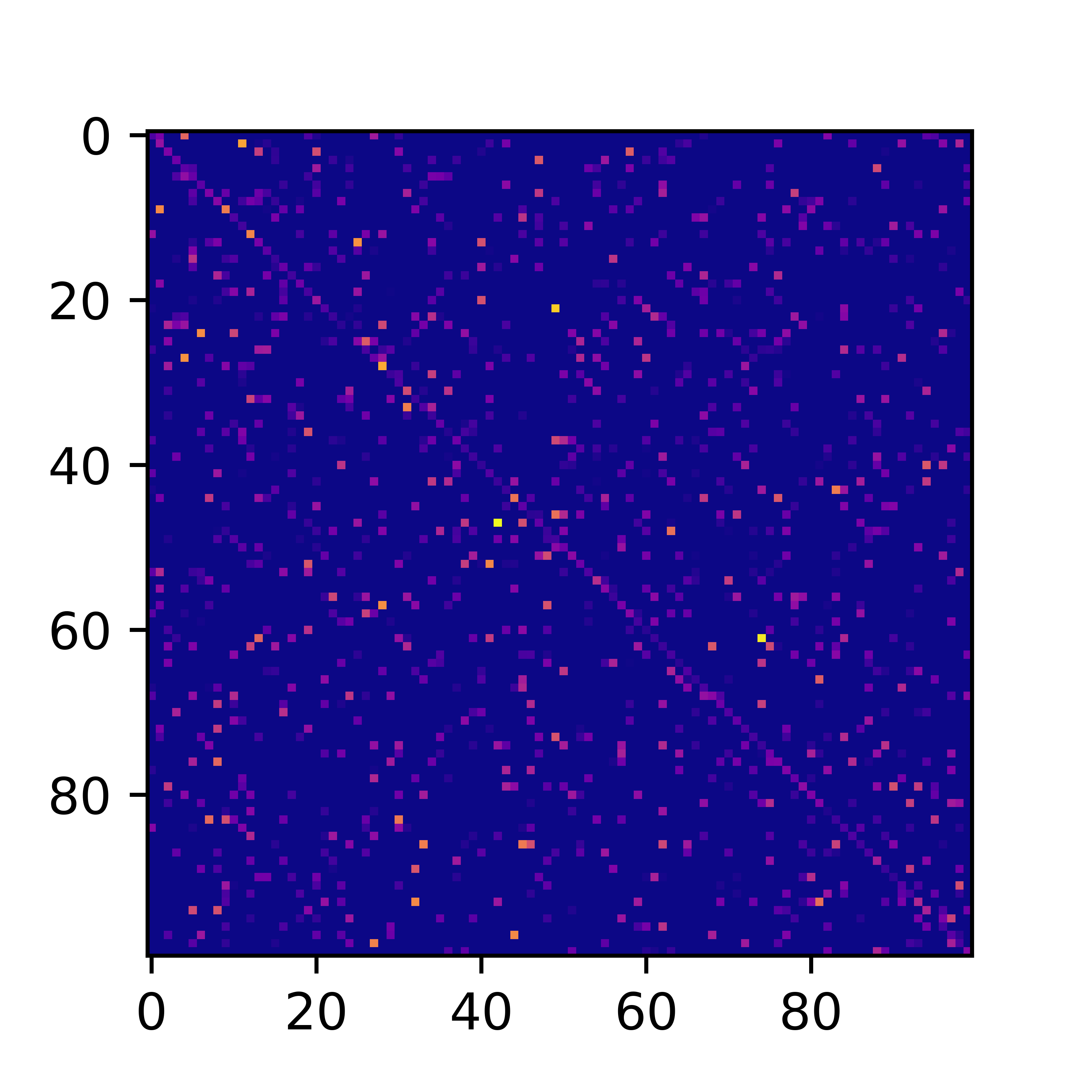}} \quad 
\caption{Examples of the learned sparsity patterns on the mixing matrices on FB15K-237} 
\label{fig:learnedpattern}
\end{figure}

The sparsest SRT on FB15K-237 that we found had only $0.13\%$ of active
entries in $\tG$ ($d_e=100$, $d_r= 100$, $d_r^*=105$), yet it performed
competitive with ComplEx with the same embedding size $100$ (27.0\% vs. 27.1\%
MRR, 45.0\% vs. 45.0\% HITS@$10$). Examples of the learned mixing matrices of
relations (chosen from top-$30$ most frequent relations) of different semantic
categories are visualized in Figure~\ref{fig:learnedpattern}. Here lighter dots
indicate stronger interactions. As can be seen, the diagonal entries play an
important role for both the symmetric relation $win\_same\_award$ and the
asymmetric relation $location\_contains$, while they are not so pronounced for
the other two relations.

To get some additional insight, we compared the performance of the
aforementioned SRT and ComplEx models on the top-$5$ and bottom-$5$ of the $30$
most frequent FB15K-237 relations, sorted by the sum of the absolute diagonal
values of SRT's mixing matrices. This relation-wise performance comparison shows
that ComplEx had better results (29.5\% vs 31.2\% MRR, 44.7\% vs 46.3\%
HITS@$10$) for relations where the learned diagonal interactions of SRT were
strong (top-$5$), while SRT performed better (19.6\% vs 18.7\% MRR, 43.0\% vs
40.7\% HITS@$10$) for relations where the learned diagonal interactions were
weak (bottom-$5$). The learned patterns for $win\_same\_award$ and
$location\_contains$ agree to some extent with the prescribed patterns of
DistMult, ComplEx, and Analogy, where the diagonal entries belong to the few
active entries allowed to be active. The relation-wise performance comparison
indicates that the prescribed constraint of ComplEx is not always suitable for all
relations.

\subsection{Computational Cost} 

Training SRT/DRT models is more expensive than training ComplEx, because the
whole core tensor is updated in every training step. In our implementation using
a single Titan X GPU, SRT/DRT with $d_e=200$ on FB15K-237 required around
$6$ hours to train for a single configuration, while ComplEx with $d_e=200$
required about $1$ hour. Generally, a key advantage of models such as ComplEx is
that they are more computationally-friendly.
% The computational cost for a single update in SRT/DRT
% % $\mathcal{O}(2d_e^2d_r^*+d_r^*+2d_e)$. It 
% is more expensive compared with ComplEx during training, as the whole core tensor needs to be updated for every training step. In contrast, only one frontal slice needs to be updated for ComplEx. 
% In our implementation on a single Titan X GPU, SRT/DRT with $d_e=200$ takes around $6$ hours to train for a single configuration,
% while ComplEx with $d_e=200$ takes about $1$ hour. Models with prescribed constraints are more computationally-friendly. 
%In prediction, the cost is comparable to the baseline models as we can pre-compute $\tG \times_3 \mR$ for speed-up. 
% The convergence behaviors of the best performed ComplEx, SRT and RT on FB15K-237 is shown in Figure~\ref{fig:converge}.

% \begin{figure}[!h]
% \centering
% \subfigure[Convergence behavior on FB15K-237]{\includegraphics[scale=0.49]{figures/converge-fb.png}} \quad
% % \subfigure[Convergence behavior on WN18RR]{\includegraphics[scale=0.49]{figures/converge-wn.png}}
% \caption{Convergence comparison in terms of epochs}
% \label{fig:converge}
% \end{figure}

\subsection{Comparison to Previous Work}\label{sec:compvsprior}

\begin{table}[h]
  \centering
  \caption{Baseline results on FB15K-237, previous work vs. our ComplEx. The \textbf{}aseline results are taken from \citet{dettmers2018conve} for entity size $d_e=100$ and
\citet{Canonical} for $d_e=2000$. 
  }
  \label{tab:FB15k237results}
    \begin{tabular}{lrrr}
      \hline\hline 
                         & Entity & Reciprocal & MRR    \\ 
    Model                & Size   & Relations  &     \\\hline  \hline
\multicolumn{4}{c}{FB15K-237}\\ \hline
    ComplEx    & 100 & Yes       &  24.7      \\ 
    ComplEx    & 2,000 & No &   35.0      \\
    Best reported  & 2,000 & Yes & 36.0     \\ \hline
    ComplEx (Ours)      & 100 & No  & 27.1    \\ 
    ComplEx (Ours) & 200 & No  &27.2  \\   
     ComplEx (Ours)      & 600 & No  & 30.0     \\ 
\hline
      \multicolumn{4}{c}{WN18RR} \\ \hline
    ComplEx  & 100 & Yes   & 44.0 \\ 
    ComplEx      & 2000 & No      & 47.0  \\ 
    Best reported    & 2000 & Yes  & 48.6   \\ \hline 
    ComplEx (Ours)  & 100 & No & 44.6   \\ 
    ComplEx (Ours)  & 200 & No & 47.0  \\
    ComplEx (Ours)  & 600 & No &   49.6 \\      
    \hline
\hline
    \end{tabular}
\end{table}

\label{sec:complexbaseline}

We conclude this section by relating the performance of our ComplEx baseline
to prior work; see Table~\ref{tab:FB15k237results}.  First note that \citet{dettmers2018conve} applied a
``reciprocal relations'' trick~\cite{Canonical}\footnote{See  supplementary material for the details} during training.
%  according to
% its published source code: The data is augmented by adding a triple
% $(j,k^{-1},i)$ for each triple $(i,k,j)$ in the training data. The model then
% learns two relation embeddings $\vr_k$ and $\vr_{k^{-1}}$. Instead of
% answering questions of form \textit{(?,   k, j)} and \textit{(i,k,?)}, the
% model answers \textit{(j, $k^{-1}$, ?)} and \textit{(i, k, ?)}. This approach
% empirically improved the prediction accuracy~\cite{Canonical} in prior
% studies. Observe that the relation sizes are now doubled, and 
This trick makes the score of a
triple $(i,k,j)$ becomes ambiguous.
 % (since generally $s(i,k,j)\neq
% s(j,k^{-1},i)$). 
We did not use this trick in our implementations.  %It is
also unclear how to use reciprocal relations in other evaluation protocols
like triple classification in a principled way. In our implementation, we did
\textit{not} use reciprocal relations.

On both FB15K-237 and WN18RR, our ComplEx achieved better results than the one
from~\citet{dettmers2018conve} with the same entity embedding size $100$.
ComplEx with $d_e=2000$ and reciprocal trick are the best performing models on FB15K-237. On WN18RR, our ComplEx  with
$d_e=600$ achieved the best results without using
reciprocal relations. These results establish that our ComplEx implementation is a strong
baseline model.

\section{Conclusion}\label{sec:conclusion}

In this study we introduced the RT decomposition, and 
showed that existing BM are special instances of RT. 
In contrast to BM, the RT decomposition allows parameters sharing across different
relations, and it decouples the entity and relation embedding sizes. A comparison of sparse RT (SRT)
and dense RT (DRT) and ComplEx suggested 
that both properties can be beneficial. 
Dense models \textit{can}  achieve competitive results.
While models such as ComplEx are more computationally-friendly, it is sometimes possible and even beneficial to learn better
sparsity patterns. 
%  asked if the sparsity of bilinear models is 
% important for their good performance and 
% found that an SRT \emph{can} perform similar or better 
% than ComplEx, indicating that Likewise, we observed that a DRT \emph{can} outperform both
% SRT and ComplEx with a similar effective number of parameters and with only a
% fraction of the entity embedding size. However, the best model generally depends on the 
% dataset and model size requirements. Generally, a key advantage of models such as ComplEx 
% is that they are more computationally-friendly, but our results suggest that even ComplEx 
% could be improved to work better for certain relations.

%By analyzing the learned sparsity patterns of a SRT model, we found that the diagonal 
%entries in the resulting mixing matrices are not pronounced in many frequent relations 
%and that SRT performed better than ComplEx on those, showing that the prescribed 
%constraints might not always be suitable for all relations. 

%%% Local Variables:
%%% mode: latex
%%% TeX-master: "main"% Table~\ref{tab:data}.
%%% End:

%%%%%%%%%%%%%%%%%%%%%%%%%%%%%%%%%%%%%%%%%%%%%%%%%%%%%%%%%%%%%%%%%%%%%%%%%%%%%%%
%%%%%%%%%%%%%%%%%%%%%%%%%%%%%%%%%%%%%%%%%%%%%%%%%%%%%%%%%%%%%%%%%%%%%%%%%%%%%%%
\clearpage
\bibliography{references}
\bibliographystyle{icml2019}

% !TEX root = main.tex
\clearpage
\section{Supplementary Materials}

\begin{table}[!h]
  \centering
  \caption{Baseline results; previous work vs. our ComplEx 
  }
  \label{tab:baseline}
    \begin{tabular}{lrrrrrrrr}
      \hline\hline
                         & Entity & Reciprocal & Effective  & Effective    & MRR  & \multicolumn{3}{c}{HITS}  \\ 
    Model                 & Size & Relations & Relation  &Number of &  & @1 & @3 & @10   \\
                   
                   &  &  & Size $d_r^*$  &  Parameters  &   &  \\  \hline  \hline
\multicolumn{9}{c}{FB15K-237}\\ \hline
    ComplEx \cite{dettmers2018conve}    & 100 & Yes & 200 & 1,547K      &  24.7   &  15.8    &  27.5      & 42.8    \\ 
 %   ComplEx \cite{Canonical}     & 2,000 & No & 2,000 & 30,474K  &  35.0   &  -         &  -       & 54.0    \\
    Best reported \cite{Canonical} & 2,000 & Yes & 2,000 &  30,948K &  36.0  &  -         &  -       & 56.0    \\ \hline
%    ComplEx (Ours)      & 100 & No & 100 &  1,524K     & 27.1  &  18.2    &  30.1      &  45.0  \\ 
    ComplEx (Ours) & 200 & No & 200 & 3,047K &27.2 &  17.6 & 30.4 & 46.7  \\   
     ComplEx (Ours)      & 600 & No & 600 &  9,142K     & 30.0    &  20.7    &  33.2      &  48.3  \\ 
\hline
      \multicolumn{9}{c}{WN18RR} \\ \hline
    ComplEx \cite{dettmers2018conve} & 100 & Yes & 100 &      4,058K        & 44.0 & 41.0 & 46.0 & 51.0 \\ 
%    ComplEx \cite{Canonical}     & 2000 & No & 2000 &      81,148K        & 47.0 & - & - & 54.0 \\ 
    Best reported \cite{Canonical}    & 2000 & Yes & 2000 & 81,170K   & 48.6     &  -         &  -       & 57.9  \\ \hline 
%    ComplEx (Ours)  & 100 & No & 100   &         4,057K     & 44.6&  41.1 & 46.6 &50.7  \\ 
    ComplEx (Ours)  & 200 & No  & 200 & 8,114K & 47.0 &  42.0 & 50.0 &55.4 \\
    ComplEx (Ours)  & 600 & No & 600   &         24,344K     &  49.6 &45.4 &51.7& 57.1 \\      
    \hline
\hline
    \end{tabular}
\end{table}

\begin{table}[!h]
   \centering
   \caption{ Hyper-parameter ranges in our experiments.   }
   \label{tab:hyperparameter}
     \begin{tabular}{lccc}\hline\hline
                              & SRT                             &   RT                              & ComplEx                        \\ \hline
     $\lambda$                & $\{5\nex1, 1\nex1, 1\nex2\}$     &    -                               & -                              \\
     $d_e$                    & $\{50, 100, 200\}$               & $\{50, 100, 200\}$                 & $\{50, 100, 200, 600\}$        \\
     $d_R$ (WN18RR)             & $\{7, 11, 14\}$                  & $\{7, 11, 14\}$                          & -        \\
     $d_R$ (FB15k-237)          & $\{100, 160, 200, 237,300\}$     & $\{100, 160, 200, 237,300\}$                          & -        \\ \hline 
                              & \multicolumn{3}{c}{SRT, RT, ComplEx}                        \\ \hline
     $\eta$                   & \multicolumn{3}{c}{$\{0.0, 0.1, 0.2, 0.3, 0.4,$ $ 0.5\}$}                                              \\ 
     $lr$                     & \multicolumn{3}{c}{$\{5\nex1, 1\nex1, 5\nex2, 1\nex2, 5\nex3, 1\nex3 \}$}                              \\
     $\omega$                 & \multicolumn{3}{c}{$\{10^{-4}, 10^{-5}, 10^{-6}, 0\}$}                                                 \\

     \hline\hline
     \end{tabular}
\end{table}

 \textbf{Detailed Base Line Results} The full results for baseline comparison is summarized in Tab.~\ref{tab:baseline}.

 \textbf{Hyper-parameter Ranges} 

Hyper-parameter ranges for dropout  $\eta$, learning rate $lr$, weight decay $\omega$, relation embedding sizes $d_R$, $l_0$ regularization parameter $\lambda$.
   For SRT we set the hyper-parameters for all hard-concrete variables used in $l_0$ regularization fixed with location mean $loc_{mean}=3.0$,  location standard deviation $loc_{std}=1$, location temperature $\beta=2/3$, stretch ranges $\zeta=1.1$ and $\gamma=-0.1$. We set $\lambda=0$ for the first $25$ epochs, i.e. we initially do not include the $l_0$ regularization term, as it empirically  improved the results. 
   
 \textbf{Reciprocal Trick}

 The data is augmented by adding a triple
 $(j,k^{-1},i)$ for each $(i,k,j)$ in the data. The model then
 learns two relation embeddings $\vr_k$ and $\vr_{k^{-1}}$. Instead of
 answering questions \textit{(?,   k, j)} and \textit{(i,k,?)}, the
 model answers \textit{(j, $k^{-1}$, ?)} and \textit{(i, k, ?)}. This approach
 empirically improved the prediction accuracy~\cite{Canonical} in prior
 studies. Observe that the relation sizes are doubled and the score of a
 triple $(i,k,j)$ becomes ambiguous (since generally $s(i,k,j)\neq s(j,k^{-1},i)$).

% %%% Local Variables:
% %%% mode: latex
% %%% TeX-master: "main"
% %%% End:
\end{document}